\documentclass[10pt,twocolumn,letterpaper]{article}

\usepackage{wacv}
\usepackage{times}
\usepackage{epsfig}
\usepackage{graphicx}
\usepackage{amsmath}
\usepackage{amssymb}
\usepackage{gensymb}
\usepackage{subcaption}

\wacvfinalcopy 

\def\wacvPaperID{****} 

\ifwacvfinal\fi
\begin{document}
\title{DeepFuse: An IMU-Aware Network for Real-Time 3D Human Pose Estimation from Multi-View Image}

\author{Fuyang Huang \hspace{1cm} Ailing Zeng \hspace{1cm} Minhao Liu \hspace{1cm} Qiuxia Lai \hspace{1cm} Qiang Xu\\
The Chinese University of Hong Kong\\
{\tt\small fyhuang,alzeng,mhliu,qxlai,qxu@cse.cuhk.edu.hk}
}


\maketitle
\ifwacvfinal\thispagestyle{empty}\fi

\begin{abstract}

In this paper, we propose a two-stage fully 3D network, namely \textbf{DeepFuse}, to estimate human pose in 3D space by fusing body-worn Inertial Measurement Unit (IMU) data and multi-view images deeply. The first stage is designed for pure vision estimation. To preserve data primitiveness of multi-view inputs, the vision stage uses multi-channel volume as data representation and 3D soft-argmax as activation layer. The second one is the IMU refinement stage which introduces an IMU-bone layer to fuse the IMU and vision data earlier at data level. without requiring a given skeleton model a priori, we can achieve a mean joint error of $28.9$mm on TotalCapture dataset and $13.4$mm on Human3.6M dataset under protocol 1, improving the SOTA result by a large margin. Finally, we discuss the effectiveness of a fully 3D network for 3D pose estimation experimentally which may benefit future research.

\end{abstract}

\vspace{-0.5cm}
\section{Introduction}
As a fundamental technique for many applications (e.g., Virtual Reality (VR), Human-Computer Interaction (HCI), and animation making), human pose estimation is a long-standing research problem and has received significant attention from both academia and industry. 

While 2D human pose estimation has been extensively studied in the literature (thanks to the availability of large manually-annotated datasets), existing 3D human pose estimation techniques still have many limitations. Marker-based vision solutions (e.g., Vicon~\cite{Vicon2014:Online}) are able to achieve high accuracy in recovering 3D human pose and position, but they require sophisticated setup for the surrounding cameras as well as carefully-calibrated markers on human body. 
Markerless vision solutions (e.g., Kinect and LeapMotion) are handier, but they can only capture human pose within a near range and fail when there is occlusion. 
Alternatively, body-worn IMUs (e.g.,  Xsens~\cite{Xsens2009:Online}) show remarkable stability and accuracy in capturing bone orientation, but they cannot tell the accurate joint positions.

Considering the pros and cons of the two types of sensors, an interesting problem is whether we could fuse the IMU data and vision data to achieve better results. One challenging issue for such fusion is that the vision input is in pixel/voxel format while the IMU input is in quaternion form. The difference in feature spaces makes it difficult to directly concatenate them into a single network. Trumble \etal~\cite{trumble2017total} simply fuse the results from the two kinds of sensors with a fully connected layer at the end of the network for regression. Such a straightforward solution does not realize the potential benefits of combining the two modalities. To tackle this problem, some optimization-based solutions try to fuse the data by introducing pre-defined skeleton lengths~\cite{Pons11,malleson2017real,von2018recovering}. These solutions, however, requires pre-defined skeleton data, and thus they cannot be well generalized for unknown subjects.

To overcome the limitations of the above fusion solutions, we propose \emph{DeepFuse}, a novel IMU-aware network that can fuse the two modalities deeply by introducing a soft-argmax layer and an IMU-bone layer in the network. DeepFuse does not require a given skeleton model a priori, and hence it can be well generalized to unknown subjects. Moreover, to make full use of the geometric-related frame data from multi-view images and preserve data primitiveness, we propose a new data representation: \emph{multi-channel volume} for multi-view representation. Finally, we propose a new data augmentation technique, namely \emph{Random Shut}, to enhance the generalization capability of our network for multi-channel volume data. 

We test our fusion solution on TotalCapture~\cite{trumble2017total} dataset, featuring synchronized camera data from 8 viewpoints, 13 body-worn IMUs and high-quality ground truth. Experimental results show that the proposed approach not only improves the estimation accuracy but also makes the two modalities of sensors mutually complementary. 
In addition, we test our vision-only network on a popular dataset Human3.6M~\cite{h36m_pami} and achieve state-of-the-art results as well. 

The main contributions of this work include:
\vspace{-5pt}
\begin{itemize}
\item We propose a new vision-IMU data fusion technique namely \emph{DeepFuse} for learning-based 3D human pose estimation, which deeply fuses data from the two kinds of sensors. Unlike previous works, the pre-defined skeleton model is not required in our method, making it well generalized to unknown users. 
\vspace{-3pt}
\item We propose a new data format namely \emph{multi-channel volume} with a corresponding data augmentation algorithm namely \emph{Random Shut} to process multi-view images. This data format is able to preserve the geometric information of cameras and data primitiveness. Ablation study shows that Random Shut is effective in enhancing model generalization capability.
\vspace{-3pt}
\item To the best of our knowledge, this is the first work that applies soft-argmax layer in a fully 3D CNN network with volumetric input for human pose estimation. Our method outperforms state-of-the-art result by a large margin and we provide a detailed analysis to show effectiveness of 3D soft-argmax with volumetric representation by conducting rigorous experiments.
\vspace{-3pt}
\end{itemize}

The remainder of this paper is organized as follows. In Section~\ref{sec:related}, we review the literature on human pose estimation. Section~\ref{sec:method} details our method. Next, we present comparative study and ablation study with state-of-the-art works in Section~\ref{sec:exp}. Limitations is discussed in Section~\ref{sec:limit}. Finally, Section~\ref{sec:conc} concludes this paper.

\begin{figure}
\begin{center}
\includegraphics[width=\linewidth]{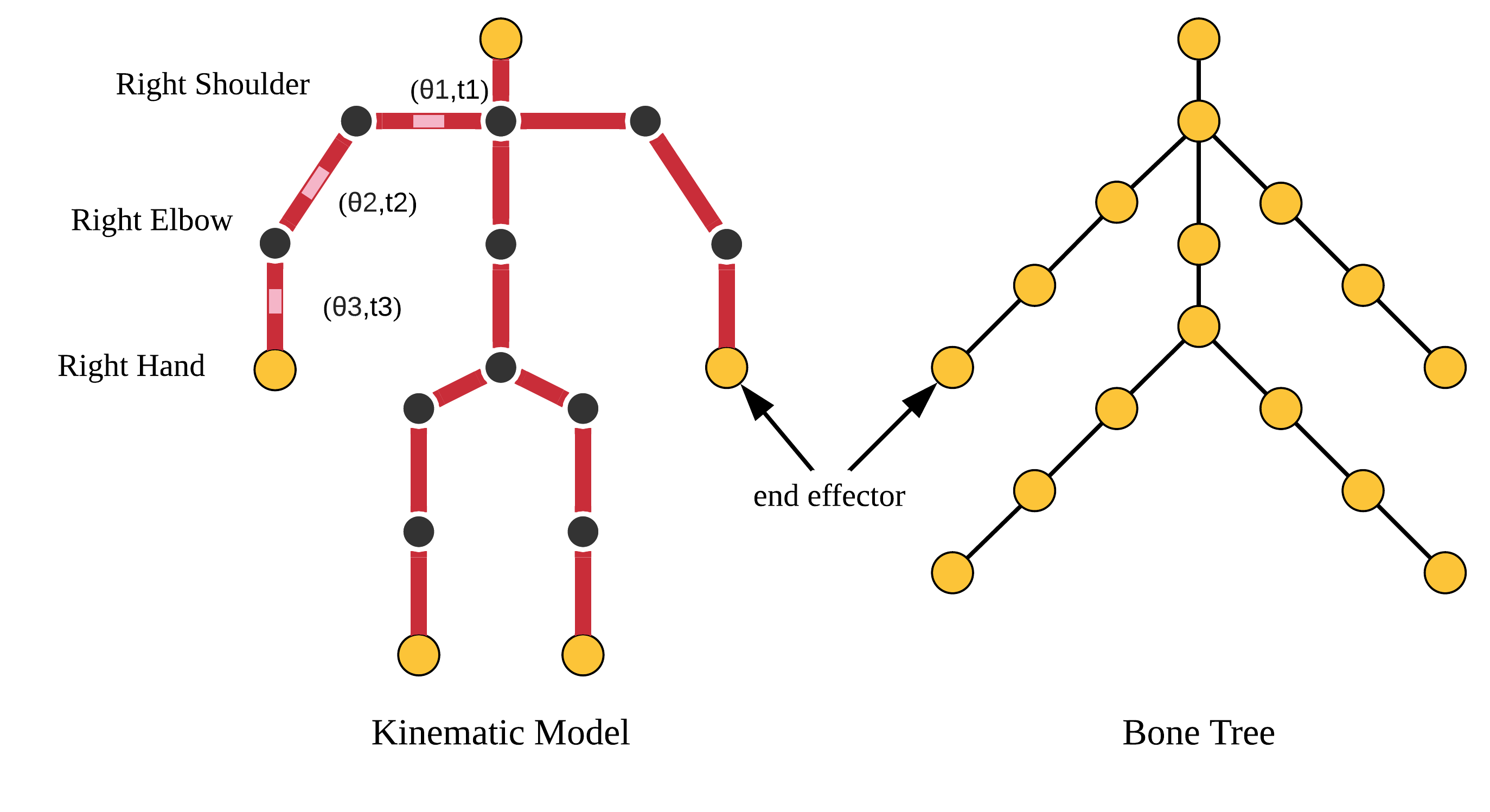}
\end{center}
    \vspace{-0.6cm}
   \caption{Illustration of kinematics. The position of end effector $P_{rh}$, right hand, can be derived through kinematic chain $P_{rh} = R(\theta_3,t_3) \cdot R(\theta_2,t_2) \cdot R(\theta_1,t_1) \cdot P_{root}$, where $R(\theta,t)$ is the rotation matrix made from IMU orientation $\theta$ and bone length $t$. See \S\ref{sec:imu} for more details.}
\label{fig:skeleton}
\vspace*{-14pt}
\end{figure}
\section{Related work}
\label{sec:related}
There are mainly three types of human pose estimation methods: vision-based, IMU-based and hybrid approaches. 
\subsection{Vision-based human pose estimation}
We can generally divide the vision-based tasks into 2D and 3D human pose estimation. 

2D human pose estimation has been extensively explored by adopting either heatmap-based approaches \cite{deepercut1,papandreou2017towards,tompson2015efficient,tompson2014joint,dantone2013human,Chen2014} or regression-based methods \cite{Ke2018,wei2016convolutional,Carreira2016} that regress 2D image to joint coordinates directly. Specifically, Newell \etal~\cite{newell2016stacked} conduct deep conv-deconv hourglass models which has been widely used as a backbone network by previous works~\cite{yang2017,bulat2016,Ke2018,chu2017,huang2018}.

Despite of the success towards 2D pose estimation, 3D pose estimation is yet under explored. Most methods for 3D pose estimation originate from 2D estimation work~\cite{li2014,yasin2016,wang2014,chen2017}. Generally speaking, heatmap-based methods \cite{luvizon20182d,mehta2017,srndi2018,sun2018integral,nibali2018,Pavlakos2016,trumble2018deep} show superior performance to that of direct regression work~\cite{lin2017,martinez2017,tekin2017,trumble2017total}. Heatmap-based methods~\cite{luvizon20182d,Pavlakos2016,sun2018integral,srndi2018} regress volumetric heatmaps from 2D image. Recently, many multi-view based methods\cite{Abdolrahim2018,tome2018rethinking,PavlakosZDD17,dong2019fast,rhodin2018learning,kocabas2019self} try to get more effective and accurate information from different views. To reduce quantization error, some work\cite{sun2018integral,luvizon20182d,nibali2018,luvizon2017human,nibali2018nu} introduce soft-argmax layer to replace hard-argmax. The effectiveness of soft-argmax on networks with 2D images input has been extensively studied in~\cite{sun2018integral}. However, the effectiveness of soft-argmax layer in a fully 3D CNN network with volumetric input for pose estimation has not been explored yet. We, therefore, conduct rigorous ablation study to demonstrate the effectiveness of soft-argmax layer in a fully 3D CNN network, shown in section~\ref{sec:sa}.

\subsection{IMU-based human pose estimation}
\label{sec:imu}
IMUs measure bone orientation accurately when they are attached to the human body. According to state-of-the-art results, the mean measurement error of bone orientation produced by body-worn IMUs is about 1.65\degree \cite{paulichxsens}, while that by the vision-based method is about 12.1\degree \cite{von2018recovering}. Consequently, they are widely used in applications wherein recovering bone orientation is sufficient~\cite{Vlasic07,paulichxsens,Tautges11,Slyper08}.

However, the IMU-based approach has several critical limitations when used to estimate human joint positions.
\begin{itemize}
    \item A pre-defined skeleton model is required to solve a kinematic chain as shown in Fig~\ref{fig:skeleton} to recover joint positions. Therefore, manual calibration of the skeleton model is mandatory for each subject. Marcard \etal~\cite{von2017sparse} combine IMUs with a skinned multi-person linear model~\cite{loper2015smpl} to recover the joint positions.
    \item Body skeleton is modeled as a tree-like structure in the kinematic model. Even if you have obtained correct limb lengths, the position of end effector node, such as hand, is determined by all its ancestor nodes, making estimation error accumulated dramatically.
    \item Last but not the least, IMU is unable to determine the position of subjects in world space. Although theoretically the subject's positions can be derived by a double integral of the acceleration data in the IMU~\cite{Tautges11,Slyper08}, such measurement error dramatically accumulates over time, making it almost impossible to calculate subject's positions.
\end{itemize}

\subsection{Hybrid approach for human pose estimation}

From the above, vision-based approaches are good at acquiring joint positions, but they are sensitive to body occlusion and illumination changes. IMU-based approaches, on the other hand, are capable of capturing accurate bone orientation stably, but fail to obtain joint positions. A hybrid solution that is able to fuse the two modalities effectively would have great potential.

Malleson \etal~\cite{malleson2017real} propose a real-time optimization approach to fuse multi-view data and IMU data by combining position term, orientation term, pose prior term, and acceleration term. Particle-based optimization is used in \cite{Pons11} to constrain orientation cues from IMU and low-dimensional manifold images cues on an inverse kinematic model. Trumble \etal~\cite{trumble2017total} propose a learning-based method to fuse volumetric data and IMU data in deep neural network. Bone orientations captured from IMU are converted to joint position by applying forward kinematics. And then, joint positions obtained from the two sources are fused at the very end of the network by fully connected layers. Consequently, the fusion layer makes limited contributions to vision tensors. Marcard \etal~\cite{von2018recovering} achieve state-of-the-art result by solving a graph-based optimization that jointly optimizes vision data and IMU data on a SMPL model. They jointly optimize their model over all frames simultaneously, making it not applicable for a real-time system. Moreover, the overwhelming majority of current fusion work use pre-defined skeleton model, restricting the generalization capability of models to unknown subjects.

In terms of sensor fusion, tight coupling (fusion in data layer) shows overall better performance than loose coupling (fusion in result layer)\cite{8259006}. We argue that for learning-based approach, the original data, instead of the estimation results, from the two data sources should be fused at the early stage of network so that the network could capture a deeper relationship of the two modalities to make them mutually complementary. Additionally, pre-defined skeleton model should not be introduced, as it would restrict the generalization capability of the model. 


\begin{figure*}
\begin{center}
\includegraphics[width=\linewidth]{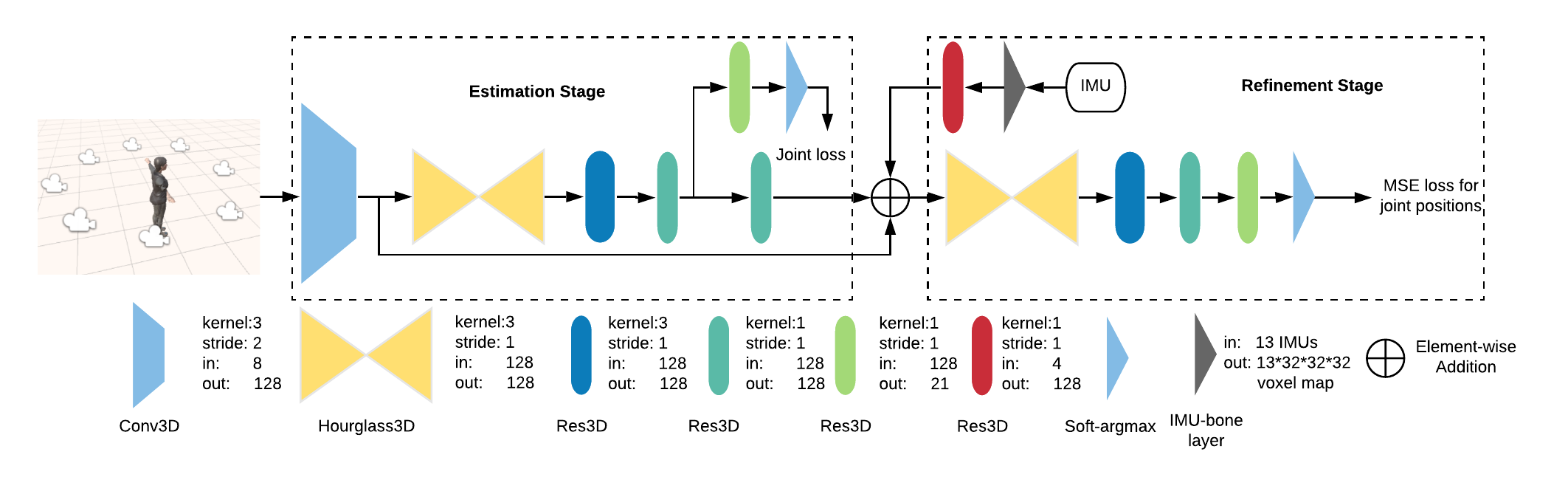}
\end{center}
  \vspace{-0.8cm}
  \caption{To simplify the illustration, all the 3D modules are visualized with 2D shapes. In the estimation stage, vision data is first down-sampled and passes through the hourglass network, Residual network 3D (Res3D) and soft-argmax layer. The first mean square error (MSE) loss between estimation result and ground truth is computed at the end of this stage. In the refinement stage, bone orientations from IMUs are transformed to $13\times32\times32\times32$ volume by IMU-bone layer, which are then concatenated with vision volume and heatmap volume. See  for more details.}
\label{fig:overall}
\vspace*{-0.2cm}
\end{figure*}

\vspace{-0.2cm}
\section{Proposed Solution}
\label{sec:method}
We define IMU-vision hybrid 3D human pose estimation as a learning-based two-stage regression problem. The network, namely \emph{DeepFuse}, takes vision and IMU data as input, and regresses 3D human joint positions directly. As shown in Fig~\ref{fig:overall}, the left part of the figure defines the estimation stage which infers 3D human pose from vision data only. The right part defines the refinement stage, introducing IMU-bone layer to refine the result from the previous stage. The whole network is trained in an end-to-end manner, making the two modalities mutually complementary
\vspace{-0.2cm}
\subsection{Pre-processing}
The vision data from TotalCapture\cite{Trumble:BMVC:2017} is captured by 8 cameras. We use the provided binary matte images as input. To make full use of the geometric correlation of multi-view images and preserve data primitiveness, we propose a new data format called multi-channel volume and an accompanied data augmentation algorithm named Random Shut. The bone orientations read from IMU are transformed from local coordination to global coordination.

\begin{figure}
    \centering
    \begin{subfigure}[b]{0.053\textwidth}
        \centering
        \includegraphics[width=\textwidth]{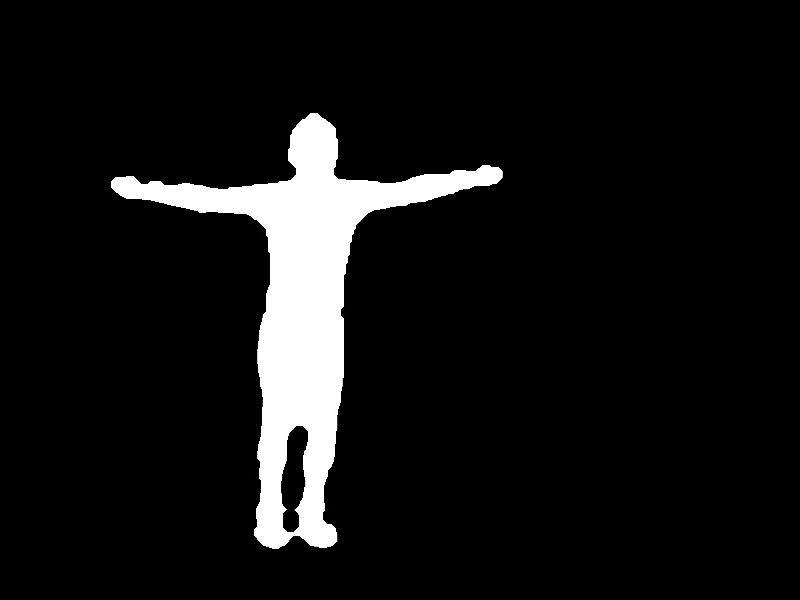}
    \end{subfigure}
   \begin{subfigure}[b]{0.053\textwidth}
   \centering
        \includegraphics[width=\textwidth]{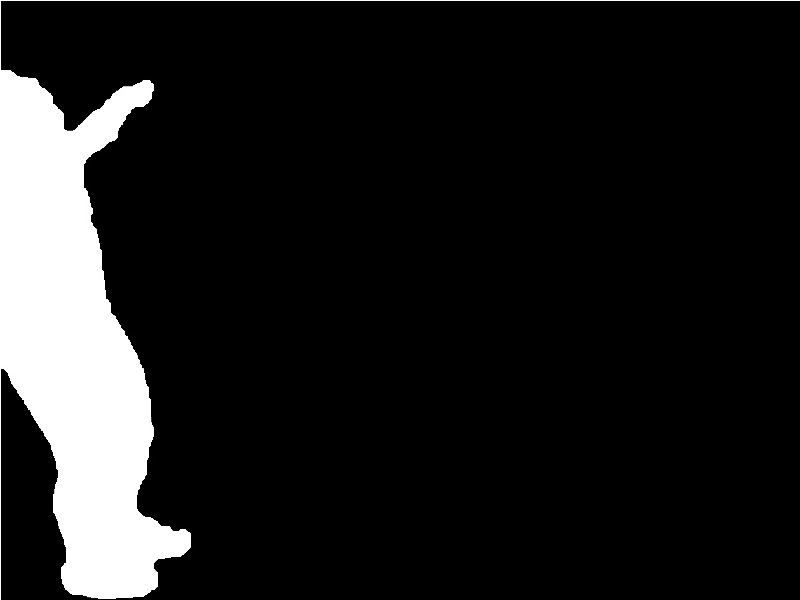}
    \end{subfigure}
    \begin{subfigure}[b]{0.053\textwidth}
    \centering
        \includegraphics[width=\textwidth]{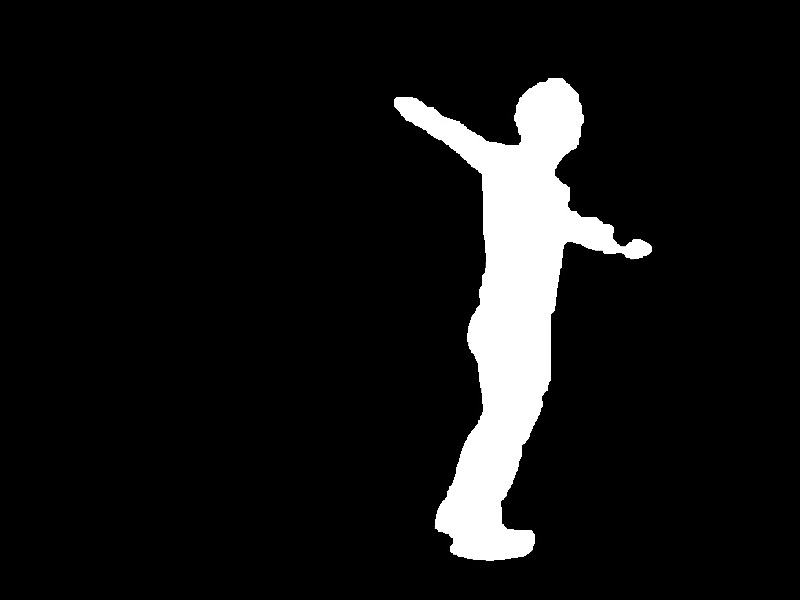}
    \end{subfigure}
    \begin{subfigure}[b]{0.053\textwidth}
    \centering
        \includegraphics[width=\textwidth]{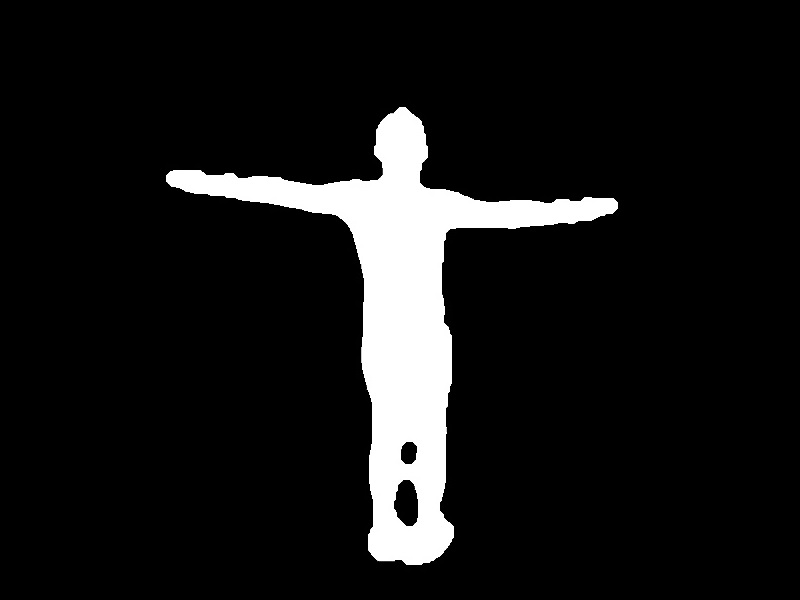}
    \end{subfigure}
    \begin{subfigure}[b]{0.053\textwidth}
    \centering
        \includegraphics[width=\textwidth]{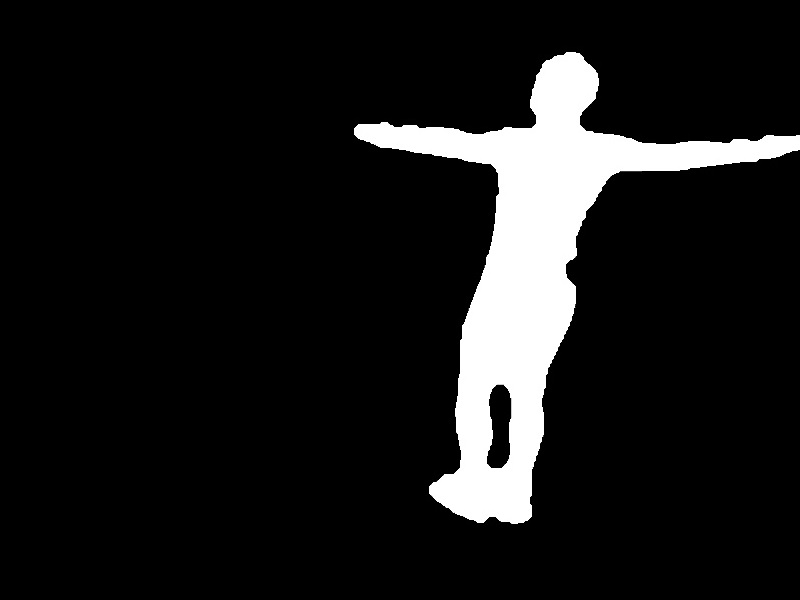}
    \end{subfigure}
    \begin{subfigure}[b]{0.053\textwidth}
    \centering
        \includegraphics[width=\textwidth]{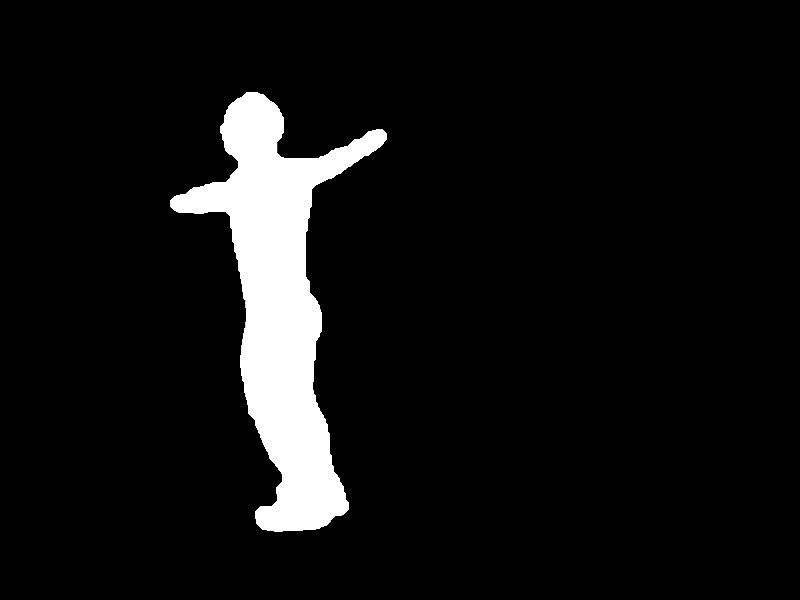}
    \end{subfigure}
    \begin{subfigure}[b]{0.053\textwidth}
    \centering
        \includegraphics[width=\textwidth]{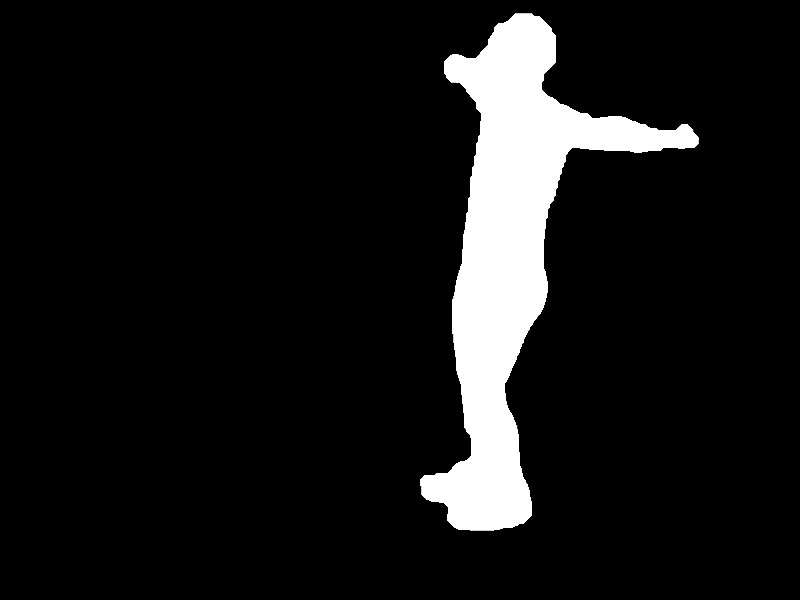}
    \end{subfigure}
    \begin{subfigure}[b]{0.053\textwidth}
    \centering
        \includegraphics[width=\textwidth]{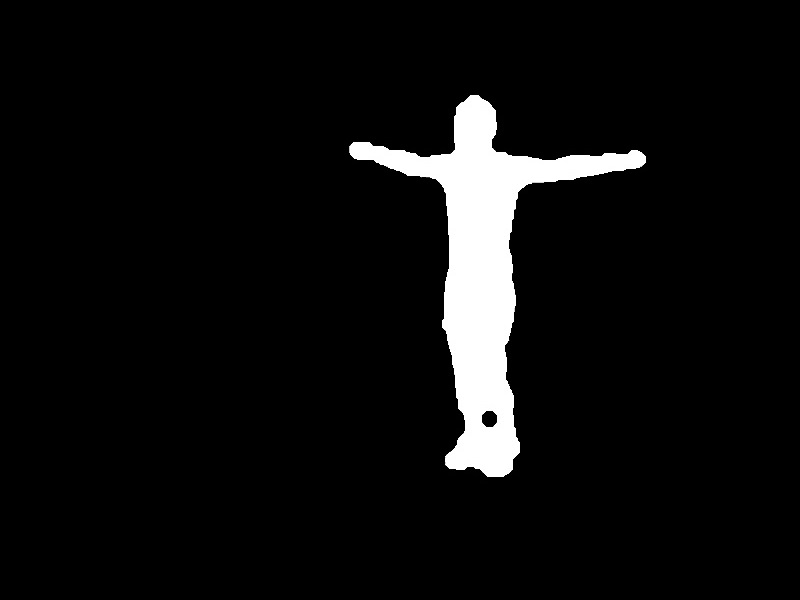}
    \end{subfigure}
    \begin{subfigure}[b]{0.053\textwidth}
        \centering
        \includegraphics[width=\textwidth]{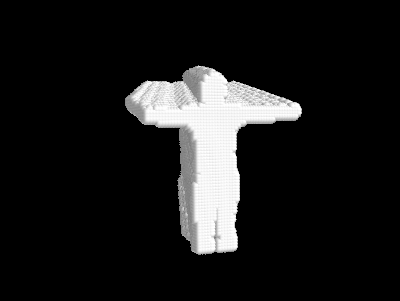}
    \end{subfigure}
   \begin{subfigure}[b]{0.053\textwidth}
   \centering
        \includegraphics[width=\textwidth]{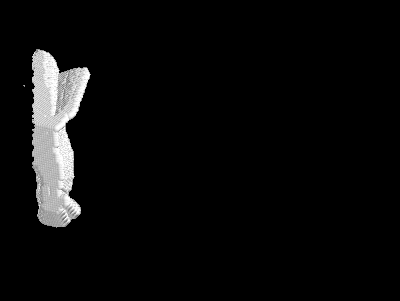}
    \end{subfigure}
    \begin{subfigure}[b]{0.053\textwidth}
    \centering
        \includegraphics[width=\textwidth]{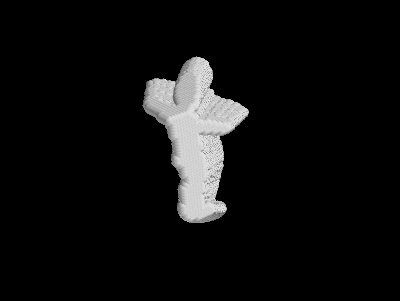}
    \end{subfigure}
    \begin{subfigure}[b]{0.053\textwidth}
    \centering
        \includegraphics[width=\textwidth]{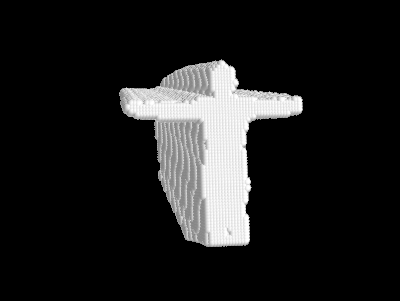}
    \end{subfigure}
    \begin{subfigure}[b]{0.053\textwidth}
    \centering
        \includegraphics[width=\textwidth]{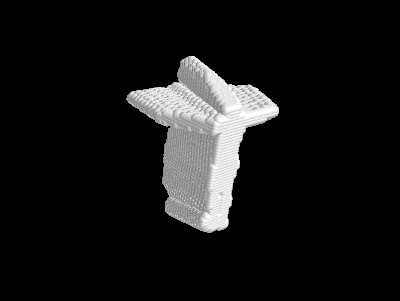}
    \end{subfigure}
    \begin{subfigure}[b]{0.053\textwidth}
    \centering
        \includegraphics[width=\textwidth]{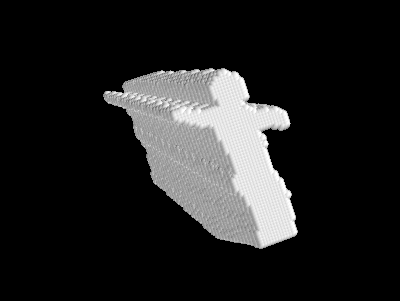}
    \end{subfigure}
    \begin{subfigure}[b]{0.053\textwidth}
    \centering
        \includegraphics[width=\textwidth]{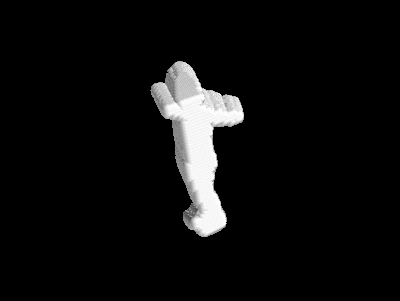}
    \end{subfigure}
    \begin{subfigure}[b]{0.053\textwidth}
    \centering
        \includegraphics[width=\textwidth]{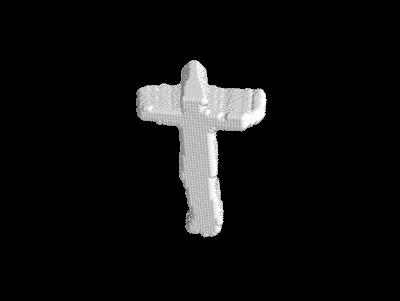}
    \end{subfigure}
    \caption{Sample multi-channel volume of a single frame. The first row is the given matte data and the second row is its respective channel of volume (See \S\ref{sec:multi_c_v}). }
    \label{fig:volume}
    \vspace*{-14pt}
\end{figure}
\vspace{-0.3cm}
\subsubsection{Multi-channel volume}
\label{sec:multi_c_v}
In terms of data representation in 3D human pose estimation, researchers either regard multi-view images as synchronized 2D images~\cite{sun2018integral,nibali2018nu} without considering multi-view geometry of cameras, or transform them into 3D volume  \cite{trumble2017total,trumble2018deep} by introducing probability visual hull (PVH). However, PVH does not well preserve data primitiveness since only one volume is generated by applying Bayes probability operator over all images. Accordingly, we propose a multi-channel volume format to overcome the limitations. 

To build multi-channel volume for a single frame, we first define binary matte images $I_1,I_2 \dots I_k$ from $K$ cameras with intrinsic parameter $M_k^{int}$ and extrinsic parameter $M_k^{ext}$. For each camera $k$, we initialize a volume $V_k$ centred on the performer with resolution $64\times64\times64$. The voxel size is set to 35mm to make sure that all body parts are within the volume. The center position $C$ of each voxel in volume $V_k$ is transformed from world coordination to pixel coordination by camera parameter $P_{x,y} = M_k^{int}\cdot M_k^{ext}\cdot C$. The voxel value is set to 1 if its corresponding pixel in image $I_k$ is occupied and to 0 otherwise. Instead of fusing the $K$ volumes into one volume by applying summation or multiplication \cite{trumble2017total,trumble2018deep} , we simply regard each volume as a single channel of input to preserve original information as much as possible. Finally, the shape of the multi-channel volume is $8\times64\times64\times64$, serving as the vision input of our network. Fig~\ref{fig:volume} shows a sample multi-channel volume and its respective matte images. 
\vspace{-0.3cm}
\subsubsection{Data augmentation}
In order to prevent overfitting, the volumes are augmented by performing a random rotation around the vertical axis within the range $[-\pi,\pi]$. Additionally, as shown in section~\ref{sec:rs}, the estimation accuracy decreases when the subject is not captured by all the 8 cameras at the same time. In order to make our model well adapted to this situation, chances are that a random volume channel is set to zero to augment training data. We set the chance of randomly 'shutting down' a camera to $20\%$ and name it as Random Shut. This data augmentation algorithm is proved to be effective for multi-view data in ablation study~\ref{sec:rs}.
\vspace{-0.3cm}
\subsubsection{IMU orientation}
According to \cite{trumble2017total}, IMUs are assumed rigid attached to human bones. The orientation data for sensor $k$ is measured in local frames and employed in quaternion representation (as in:\cite{gebre2004design,choukroun2003novel,bachmann1999orientation}), noted as $_{}^{local}\textrm{q}_{k}$. By multiplying wearing offset  $_{}^{wear}\textrm{q}_{k}$ and local-global transform quaternion $_{local}^{global}\textrm{q}_{k}$ , the bone orientation in global frame $_{}^{global}\textrm{q}_{k}$  is calculated as:
\begin{equation}
_{}^{global}\textrm{q}_{k} =   _{}^{local}\textrm{q}_{k} \otimes _{local}^{global}\textrm{q}_{k} 
\end{equation}
\begin{equation}
_{}^{global}\widehat{\textrm{q}_{k}} = _{}^{wear}\textrm{q}_{k}^{*} \otimes  _{}^{global}\textrm{q}_{k}
\end{equation}
where $"*"$ denote the quaternion conjugate.

The bone orientations in global frame $_{}^{global}\widehat{\textrm{q}_{k}}$ are then transformed into IMU-bone layer in the network as discussed in section~\ref{sec:ib}.

\begin{table}
\begin{center}
\resizebox{0.45\textwidth}{!}{
\setlength\tabcolsep{5pt}
\renewcommand\arraystretch{1.0}
\begin{tabular}{|l|c|}
\hline
Method & Error (mm)  \\
\hline\hline
Video Inertial Poser (VIP)~\cite{von2018recovering} & 26.0$\dagger$  \\
\hline
Frame-by-Frame Optimization ~\cite{malleson2017real} & 62.0\\

FC IMU+3D PVH ~\cite{trumble2017total}& 70.0\\
\emph{DeepFuse} & \textbf{28.9} \\

\hline
\end{tabular}
}
\end{center}
\vspace{-0.3cm}
\caption{Comparison results regarding mean joint error on TotalCapture dataset. 
$\dagger$ indicates sequence-based work. See \S\ref{sec:tc_eval} for detail. (Best in \textbf{bold}; same for other tables).}
\vspace{-0.4cm}
\label{tab:final}
\end{table}

\begin{table*}
\begin{center}
\resizebox{\textwidth}{22mm}{
\begin{tabular}{ l c c c c c c c c c c c c c c c c c}
\hline
\textbf{Protocol 1} & Direct & Discuss & Eat & Greet & Phone& Photo & Pose  & Purcha. & Sit & SitD &Smoke &Wait &WalkD&Walk& WalkT & Avg.\\
\hline
Fang2018 w/PA\cite{fang2018learning} & 38.2 &41.7 &43.7 &44.9& 48.5& 55.3 &40.2 &38.2& 54.5& 64.4 &47.2 &44.3 &47.3 &36.7 &41.7 & 45.7 \\ 
Sun2018  w/PA\cite{sun2018integral}  & 40.9 & 41.4 &45.0 &45.2 &42.1 &37.6 &41.1 &52.0 &71.4 &42.5 &47.4& 41.6 &32.0 &42.6 &36.9 &44.1  \\
Chen2019  w/PA\cite{chen2019weakly} &36.9 &39.3& 40.5 &41.2& 42.0 &\textbf{34.9} &38.0 &51.2 &67.5 &\textbf{42.1} &42.5& 37.5 &30.6 &40.2 &34.2 &41.6\\
Multi-View Martinez2017 w/PA\cite{martinez2017}$\times$ &39.5& 43.2& 46.4& 47.0 &51.0 &56.0 &41.4 &40.6 &56.5 &69.4 &49.2 &45.0 &49.5 &38.0 &43.1 &47.7\\
Tome2018 w/PA\cite{tome2018rethinking}$\times$ &38.2 &40.2 &38.8& 41.7& 44.5& 54.9 &34.8 &35.0 &52.9 &75.7& 43.3& 46.3 &44.7 &35.7 &37.5 &44.6\\

\emph{Ours} w/o PA $\dagger$ $\times$& \textbf{18.7}&\textbf{20.7}&\textbf{22.5}&\textbf{24.5}&\textbf{28.3}&40.1&\textbf{22.7}&\textbf{23.1}&\textbf{26.0}&\textbf{39.9}&\textbf{33.8}&\textbf{22.9}&\textbf{35.0}& \textbf{20.9} &\textbf{21.3}&\textbf{26.9}  \\
\emph{Ours} w/PA $\dagger$ $\times$& \textbf{9.0}&\textbf{9.4}&\textbf{11.2}&\textbf{13.5}&\textbf{14.0}&\textbf{22.1}&\textbf{11.6}&\textbf{12.0}&\textbf{11.8}&\textbf{20.8}&\textbf{14.7}&\textbf{10.6}&\textbf{20.4}&\textbf{10.4}& \textbf{10.7} &\textbf{13.4}  \\
\hline
\textbf{Protocol 2} & Direct & Discuss & Eat & Greet & Phone& Photo & Pose  & Purcha. & Sit & SitD &Smoke &Wait &WalkD&Walk& WalkT & Avg.\\
\hline
Trumble2017  \cite{trumble2017total} $\dagger$& 92.7 & 85.9 & 72.3 & 93.2 & 86.2 & 101.2 & 75.1 & 78.0 & 83.5&94.8 & 85.8& 82.0 & 114.6 & 94.9 & 79.7& 87.3 \\ 
Trumble2018   \cite{trumble2018deep}  $\dagger$$\times$& 41.7 & 43.2 & 52.9 & 70.0 & 64.9 & 83.0 & 57.2 & 63.5 & 61.0 & 95.0 & 70.0 & 62.3 & 66.2 & 53.7 & 52.4 & 62.5  \\
Multi-View Martinez2017 \cite{martinez2017}\cite{tome2018rethinking}$\times$ &46.5 &48.6 &54.0 &51.5 &67.5 &70.7 &48.5 &49.1 &69.8 &79.4 &57.8 &53.1 &56.7 &42.2 &45.4 &57.0 \\
Pavlakos2017\cite{PavlakosZDD17}$\times$ & 41.2 &49.2 &42.8 &43.4 &55.6 &46.9 &40.3 &63.7 &97.6 &119.0 &52.1 &42.7 &51.9 &41.8 &39.4 &56.9\\
Tome2018 \cite{tome2018rethinking}$\times$ &43.3 &49.6 &42.0 &48.8 &51.1 &64.3 &40.3 &43.3 &66.0 &95.2 &50.2 &52.2 &51.1 &43.9 &45.3 &52.8\\
Kocabas2019-FS\cite{kocabas2019self}$\times$&-&-&-&-&-&-&-&-&-&-&-&-&-&-&-&51.8\\
Kadkhodamohammadi2018\cite{Abdolrahim2018}$\times$ &39.4 &46.9 &41.0 &42.7& 53.6 &54.8 &41.4& 50.0 &59.9 &78.8 &49.8 &46.2 &51.1 &40.5 &41.0 &49.1\\
\emph{Ours} $\dagger$ $\times$ &\textbf{26.8} &\textbf{32.0} &\textbf{25.6} &52.1 &\textbf{33.3} &\textbf{42.3} &\textbf{25.8}& \textbf{25.9} &\textbf{40.5} &76.6 &\textbf{39.1} &54.5 &\textbf{35.9} &\textbf{25.1} &\textbf{24.2} &\textbf{37.5}\\

\hline
\end{tabular}}
\end{center}
\caption{Comparison results regarding mean joint error following protocol 1 and 2 of Human3.6M with 17 keypoints. $\dagger$ use provided matte data. $\times$ use multi-view images. (FS: fully supervised baseline). See \S\ref{sec:human_36} for details.}
\label{tab:human}
\vspace*{-14pt}
\end{table*}

\subsection{Network structure}
As shown in Fig~\ref{fig:overall}, DeepFuse consists of an estimation stage and a refinement stage. The backbone network used is Hourglass Network~\cite{newell2016stacked}. The network takes multi-channel volume as vision input, so we modify original Hourglass Network to the 3D version with 3D CNN. The left part of Fig~\ref{fig:overall} represents the estimation stage which takes vision data only as input and outputs 3D voxel heatmaps of each joint. Considering the large GPU memory consumption of 3D CNN, the volume input resolution for one channel is $64\times64\times64$ and the voxel heatmap is $32\times32\times32$. By adding a soft-argmax layer to the end of the hourglass network, the 3D positions of each joint can be directly regressed from voxel heatmap. Soft-argmax is differentiable so that the estimated positions are able to further propagate to generate IMU-bone layer. Thus, the entire network can be trained in an end-to-end manner.

The right part is the refinement stage, taking IMU data and output from the previous stage as input. Specifically, the IMU data and estimated joint positions from the last stage consist the IMU-bone layers which turn quaternion data into a multi-channel volume. Thus, IMU-bones layers, voxel heatmap and original vision data, can be concatenated in the same feature space. In the following stack of hourglass network, IMU data and vision data are fused implicitly to produce a set of refined voxel heatmaps. A soft-argmax layer is appended to the last layer as well, from which the final refined joint positions are estimated.  
\vspace{-0.4cm}
\subsubsection{Soft-argmax layer }
\label{sec:sal}
Heatmap-based approaches are proved to be effective in both human pose and hand pose estimation~\cite{newell2016stacked,huang2018}. The voxel heatmap is discrete, while the joint positions are continuous in world coordination. So if we assume that volume length is 2000mm and heatmap resolution is $32\times32\times32$, the average estimation error within a single voxel is about $2000/32/2=31.25mm$ at least. So even if we obtained a perfect voxel heatmap, the accuracy of joint positions in world coordination would be far from satisfactory by simply picking the largest voxel in a voxel heatmap. 

To overcome the limitation of low resolution of voxel heatmap, we introduce soft-argmax layer into our network. Instead of simply picking the largest voxel as joint position, the soft-argmax layer is able to learn a weighted average of multiple voxels to predict joint positions. 

Specifically, soft-argmax shares similar idea with \textit{softmax} algorithm. Softmax value of voxel $x_i$ is defined as:
\begin{equation}
\label{equa:sm}
    f_{softmax}(x_i)=\frac{e^{\theta x_i}}{\sum_j^Ne^{\theta x_j}}
\end{equation}
The sum of softmax values equals to 1. Thus, the coordinate of max value among all voxels is the sum of softmax values multiplied by indices along each axis: 
\begin{equation}
    \label{equa:sam}
     f(x)=\sum_i\frac{e^{\theta x_i}}{\sum_j^Ne^{\theta x_j}}i
\end{equation}
where $i$ is the x-,y-,z-coordinate of voxel $x_i$ for each axis and $N$ is the total voxel number in the volume. The larger $\theta$ will enlarge the voxels with big value and lower the smaller ones, which means if $\theta$ is large enough, soft-argmax will return the coordinate of voxel with maximum value. However, we expect that the joint position should be the coordinate of the weighted average of several large voxels. 
We set $\theta=3$ empirically (see ablation study in section~\ref{sec:sa}).

\vspace{-0.4cm}
\subsubsection{IMU-bone layer and sensor fusion}
\label{sec:ib}
As explained in section~\ref{sec:imu}, IMU sensor measures bone orientation only. Thus, the pre-defined skeleton model was introduced to estimate joint positions as ~\cite{von2018recovering,trumble2017total,von2016human} did in their work. We claim that the pre-defined skeleton model should not be used as prior knowledge in order to make the estimation model well generalized to unknown subjects. Therefore, the challenge of sensor fusing lies in that how to use measured bone orientation from IMU to refine joint position estimation result from vision. 

Existing learning-based fusion work~\cite{trumble2017total} tries to fuse the estimation results from two modalities instead of origin inputs through a dense layer at the end of the network. As a result, the original data from two sensors does not propagate to its counterpart and fusion is relatively shallow. So we are going to fuse the original input of two modalities at the early stage of the network for a deeper fusion.

According to the skeleton model, one limb $L$ consists of two joints $[J_1,J_2]$ and one piece of bone. If the position of one joint, say $J_1$, and bone orientation $q_{b}$ are given, a ray $f_{ray}(J_1,q_{b})$ can be cast from $J_1$ along $q_{b}$ direction. Thus, $J_2$ must be located somewhere along the ray. So if we can obtain the estimated joint positions $J_{est}$ from estimation stage and the bone orientation $q_b$ from IMU, a cluster of rays can be obtained respectively to describe the possible locations of joints. Since $J_{est}$ is the estimated result, the generated rays are not 100\% accurate. The rays are transformed to directed cylinders by introducing a radius $r$ around the rays. Finally, we define a channel of volume $V_i=f_{cylinder}(J_{est}[i],q_b[i],r)$ for each IMU $i$. The voxels are set to 1 if occupied by the 'bone cylinder' and to 0 otherwise. Finally, we stack these volumes together to form IMU-bone layers.   

Since we have volumetric representation for original vision data, voxel heatmap for each joint, and volumes for IMU-bone layer at this stage, the three batches of volumes can be concatenated together and then passed to next stage to refine the result from estimation stage. In this way, the original data from two modalities are deeply fused by a new stack of hourglass network in the refinement stage. Similarly, a soft-argmax layer is appended to the last layer to make final estimation of joint positions. 


\begin{figure}
\begin{center}
\includegraphics[width=\linewidth]{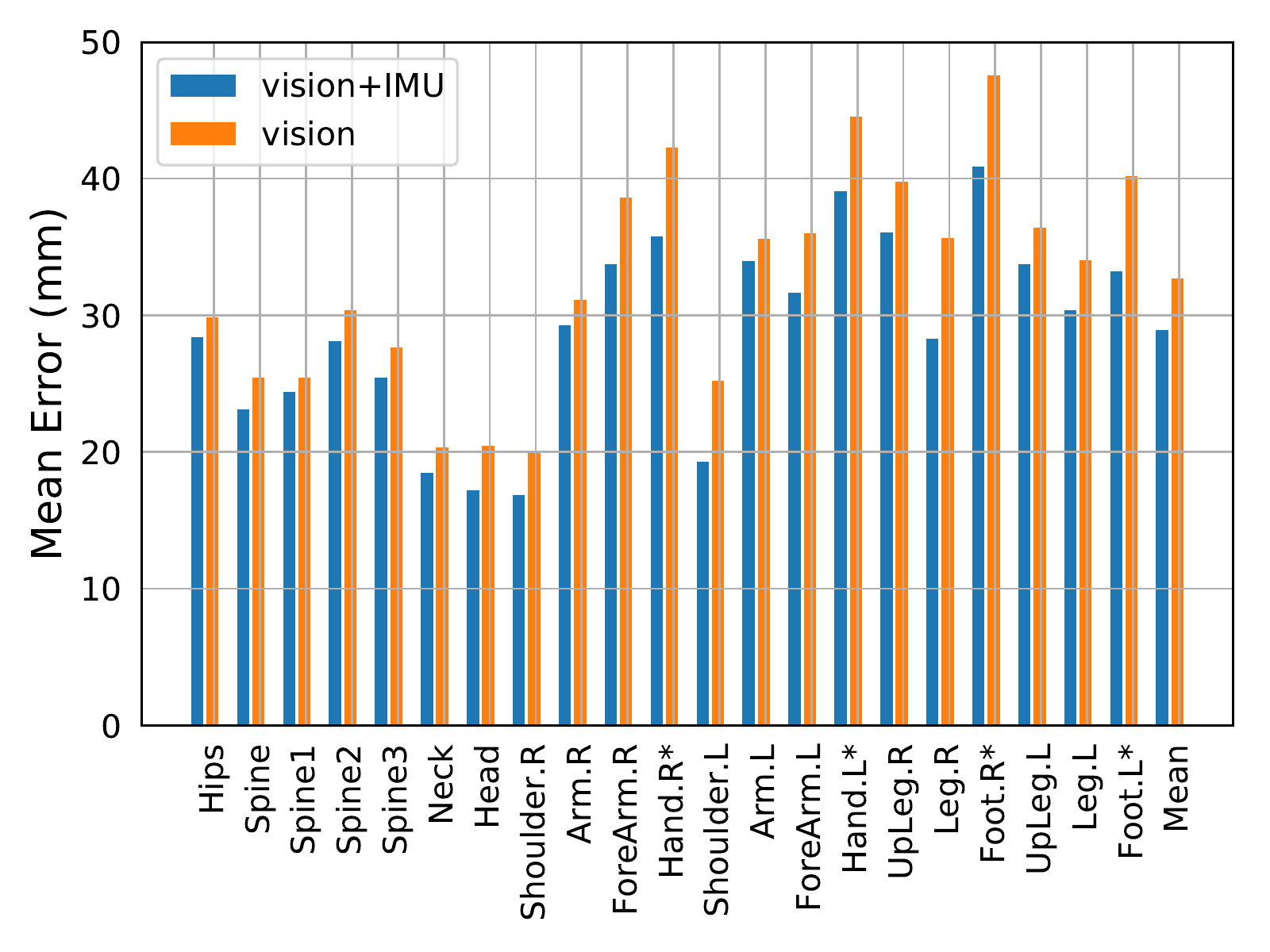}
\end{center}
    \vspace{-0.5cm}
   \caption{Per-joint mean error on test set. * indicates right/left foot and hand joints. See \S\ref{sec:fusion} for details.}
\label{fig:pj}
\end{figure}
\begin{table}
\begin{center}
\begin{tabular}{|l|c|}
\hline
Methods &  Error (mm) \\
\hline\hline
Vision-IMU w/ RS (proposed)& \textbf{28.9} \\
Vision-only w/ RS  & $32.7$ \\
\hline
Vision-IMU w/o RS  & \textbf{32.4}\\
Vision-only w/o RS & $35.9$ \\
\hline
\end{tabular}
\end{center}
\vspace{-0.5cm}
\caption{Mean joint error results for 4 different experiment settings. RS is short for Random Shut. Vision-IMU w/ RS is identical to the proposed \emph{DeepFuse} (See \S\ref{sec:fusion}). }
\label{tab:fs}
\vspace*{-14pt}
\end{table}

\vspace{-0.2cm}
\subsubsection{Training target}
By introducing soft-argmax layer, joint positions can be directly recovered. So there is no need to involve in heatmap loss as the voxel heatmap can be learned implicitly by the mean squared error between the estimated joint positions and ground truth positions in world coordination. Specifically, the loss function $L$ is the sum of estimation stage loss $L_{est}$ and refinement stage loss $L_{ref}$: 
\begin{equation}
\label{equa:loss}
  \begin{aligned}
  L&=L_{est}+L_{ref}\\&=\sum_{i}^{K}|J_{est}^{i}-J_{gt}^{i}|^{2}+\sum_{i}^{K}|J_{ref}^{i}-J_{gt}^{i}|^{2} 
  \end{aligned}
\end{equation}
where $J_{est}^{i}$ and $J_{ref}^{i}$ are the estimated joint positions for $i^{th}$ joint in estimation stage and refinement stage, respectively. $J_{gt}^{i}$ are the ground truth. K is the \#joints to be estimated. 

\section{Experimental Results}
\label{sec:exp}
 
In our experiments, RMSProp optimizer is used for training. The learning rate is initially set to be 1e-5 and decays 0.2 every 5 epochs. Our system is able to run at 25 Hz with a single NVIDIA 1080Ti GPU.

\begin{figure}
    \centering
    \begin{subfigure}[b]{0.11\textwidth}
        \centering
        \includegraphics[width=\textwidth]{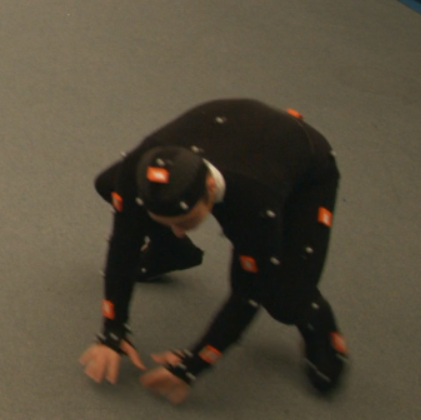}
    \end{subfigure}
   \begin{subfigure}[b]{0.11\textwidth}
   \centering
        \includegraphics[width=\textwidth]{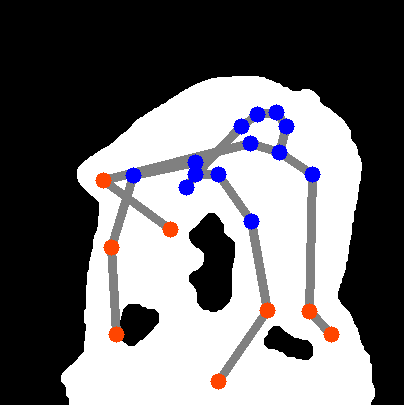}
    \end{subfigure}
    \begin{subfigure}[b]{0.11\textwidth}
    \centering
        \includegraphics[width=\textwidth]{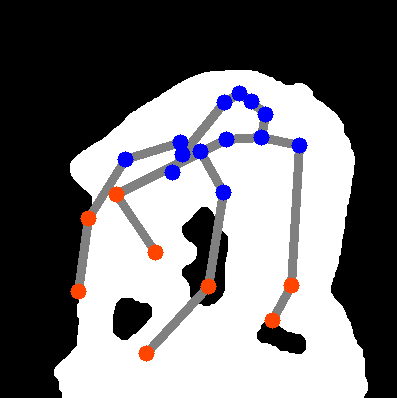}
    \end{subfigure}
    \begin{subfigure}[b]{0.11\textwidth}
    \centering
        \includegraphics[width=\textwidth]{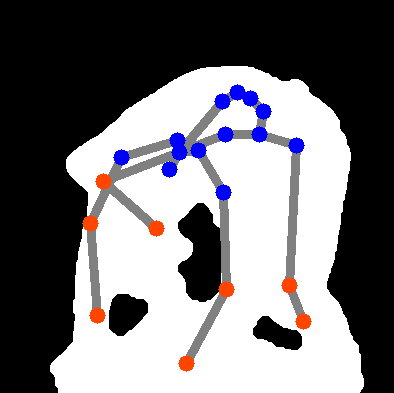}
    \end{subfigure}
    \centering
    \begin{subfigure}[b]{0.11\textwidth}
        \centering
        \includegraphics[width=\textwidth]{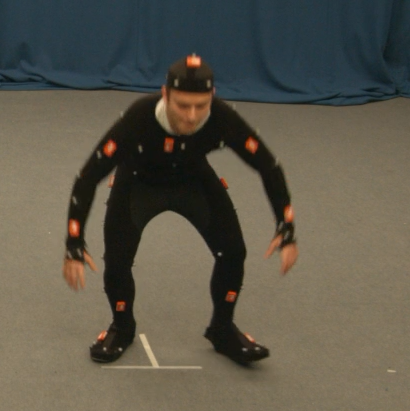}
    \end{subfigure}
   \begin{subfigure}[b]{0.11\textwidth}
   \centering
        \includegraphics[width=\textwidth]{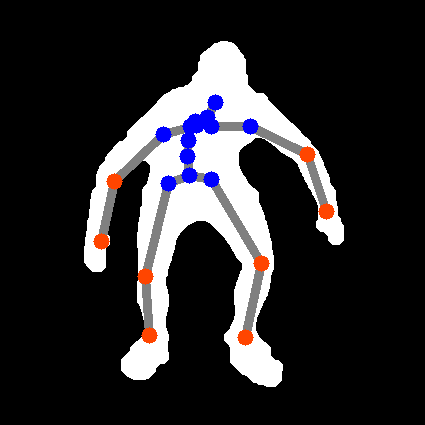}
    \end{subfigure}
    \begin{subfigure}[b]{0.11\textwidth}
    \centering
        \includegraphics[width=\textwidth]{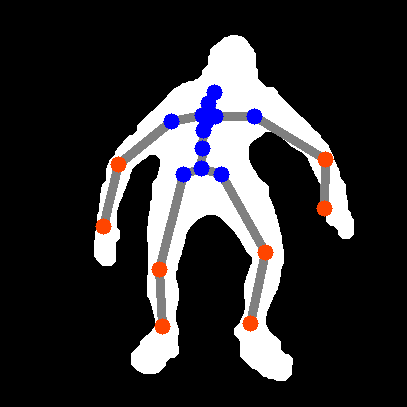}
    \end{subfigure}
    \begin{subfigure}[b]{0.11\textwidth}
    \centering
        \includegraphics[width=\textwidth]{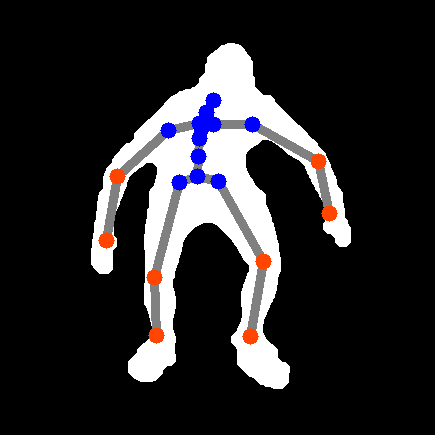}
    \end{subfigure}
    \centering
    \begin{subfigure}[b]{0.11\textwidth}
        \centering
        \includegraphics[width=\textwidth]{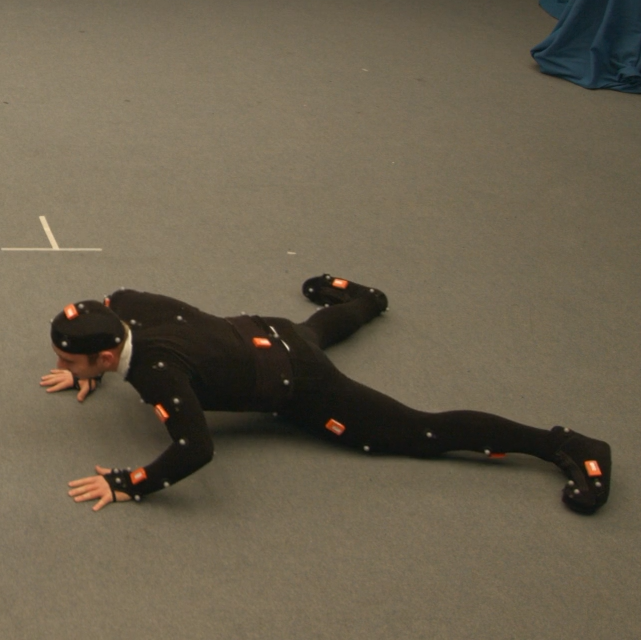}
    \end{subfigure}
   \begin{subfigure}[b]{0.11\textwidth}
   \centering
        \includegraphics[width=\textwidth]{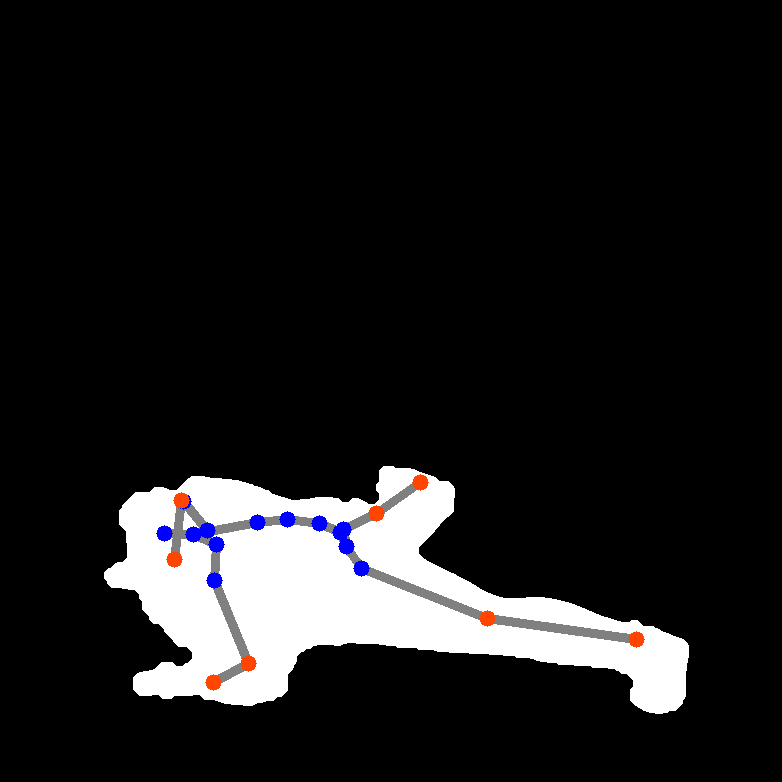}
    \end{subfigure}
    \begin{subfigure}[b]{0.11\textwidth}
    \centering
        \includegraphics[width=\textwidth]{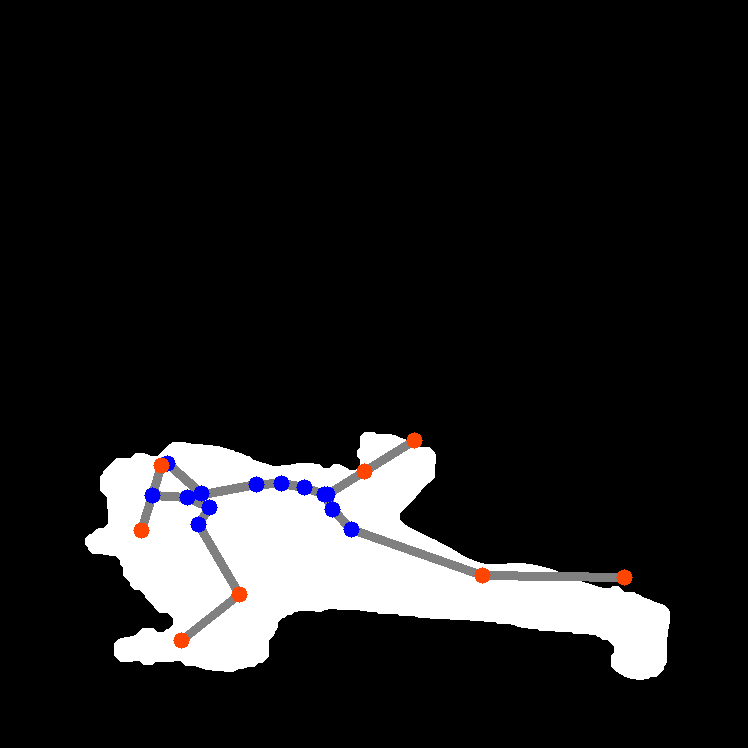}
    \end{subfigure}
    \begin{subfigure}[b]{0.11\textwidth}
    \centering
        \includegraphics[width=\textwidth]{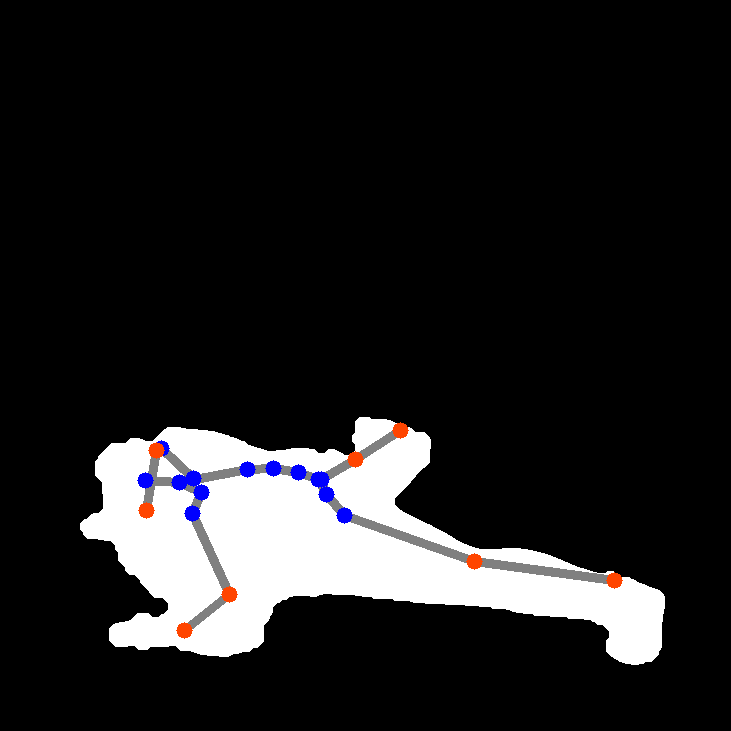}
    \end{subfigure}
    \caption{Left to right: camera, ground truth, vision-only estimation, fusion estimation. See \S\ref{sec:fusion} for details.}
    \label{fig:qual}
    \vspace*{-8pt}
\end{figure}

\subsection{Comparative study}

In this section, we present the comparative result of \emph{DeepFuse} with recent works on two public 3D human pose estimation datasets: TotalCapure~\cite{trumble2017total} and Human3.6M \cite{h36m_pami}. The TotalCapture dataset features 1.9M video frames from 8 cameras performed by 5 subjects and synchronized orientation data from 13 body-worn IMUs. Human3.6M is a more popular 3D human pose estimation dataset with only vision data as input. 

\vspace{-0.3cm}
\subsubsection{TotalCapture evaluation}
\label{sec:tc_eval}
TotalCapture~\cite{trumble2017total} is the only dataset including synchronized body-worn IMU data and multi-view video frames with high-quality ground truth. There are three pieces of work ~\cite{trumble2017total,malleson2017real,von2018recovering} evaluated on this dataset by using both vision and IMU data. Specifically, Malleson \etal~\cite{malleson2017real} optimizes the two modalities in a frame-by-frame manner while Marcard \etal~\cite{von2018recovering} optimizes the whole video sequence simultaneously. Learning-based method~\cite{trumble2017total} uses fully connected layer to fuse the IMU data and 3D vision data. 

As shown in Table~\ref{tab:final}, DeepFuse outperforms ~\cite{trumble2017total,malleson2017real} by a large margin and shows close performance with~\cite{von2018recovering}. However, pre-defined skeleton model is not employed in our method, but it was used by all the other three methods. Consequently, DeepFuse is more friendly to unknown subjects. Moreover, Video Inertial Poser (VIP)~\cite{von2018recovering} achieved the lowest estimation error by optimizing over all frames of a given video sequence and hence it is not applicable for real-time estimation. DeepFuse and the other two methods are frame-based estimations without such limitation. 

To sum up, our method achieves the state-of-the-art result in terms of real-time frame-based 3D human pose estimation by fusing IMU and vision data on TotalCapture, showing that even if there is no pre-defined skeleton model as input, vision data and IMU data can still be fused in a complementary way to produce better fusion result.

\begin{figure}
\begin{center}
\centering
    \begin{subfigure}[b]{\linewidth}
    \centering
        \includegraphics[width=\linewidth]{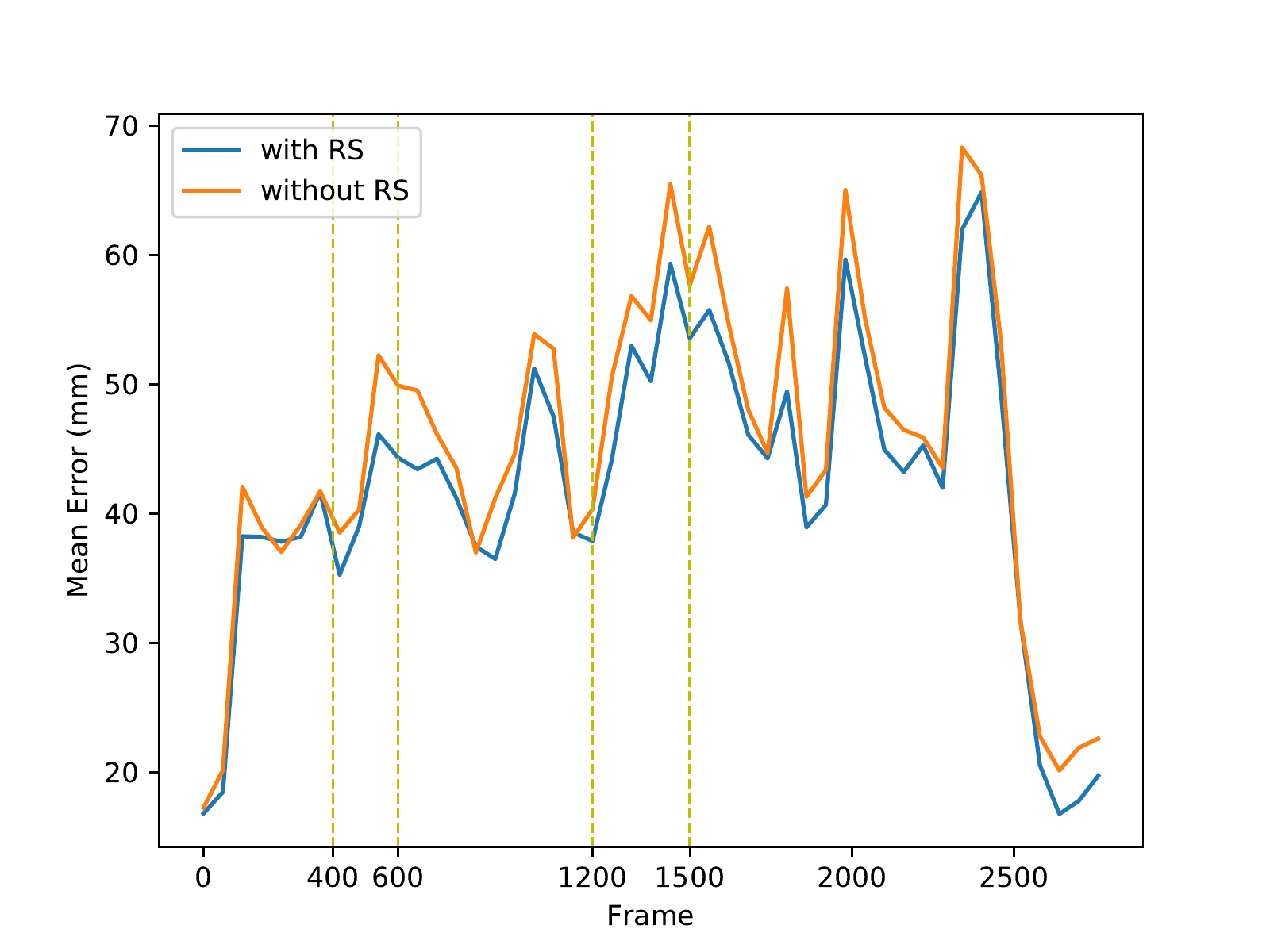}
    \end{subfigure}
    \begin{subfigure}[b]{0.1\textwidth}
    \centering
        \includegraphics[width=\linewidth]{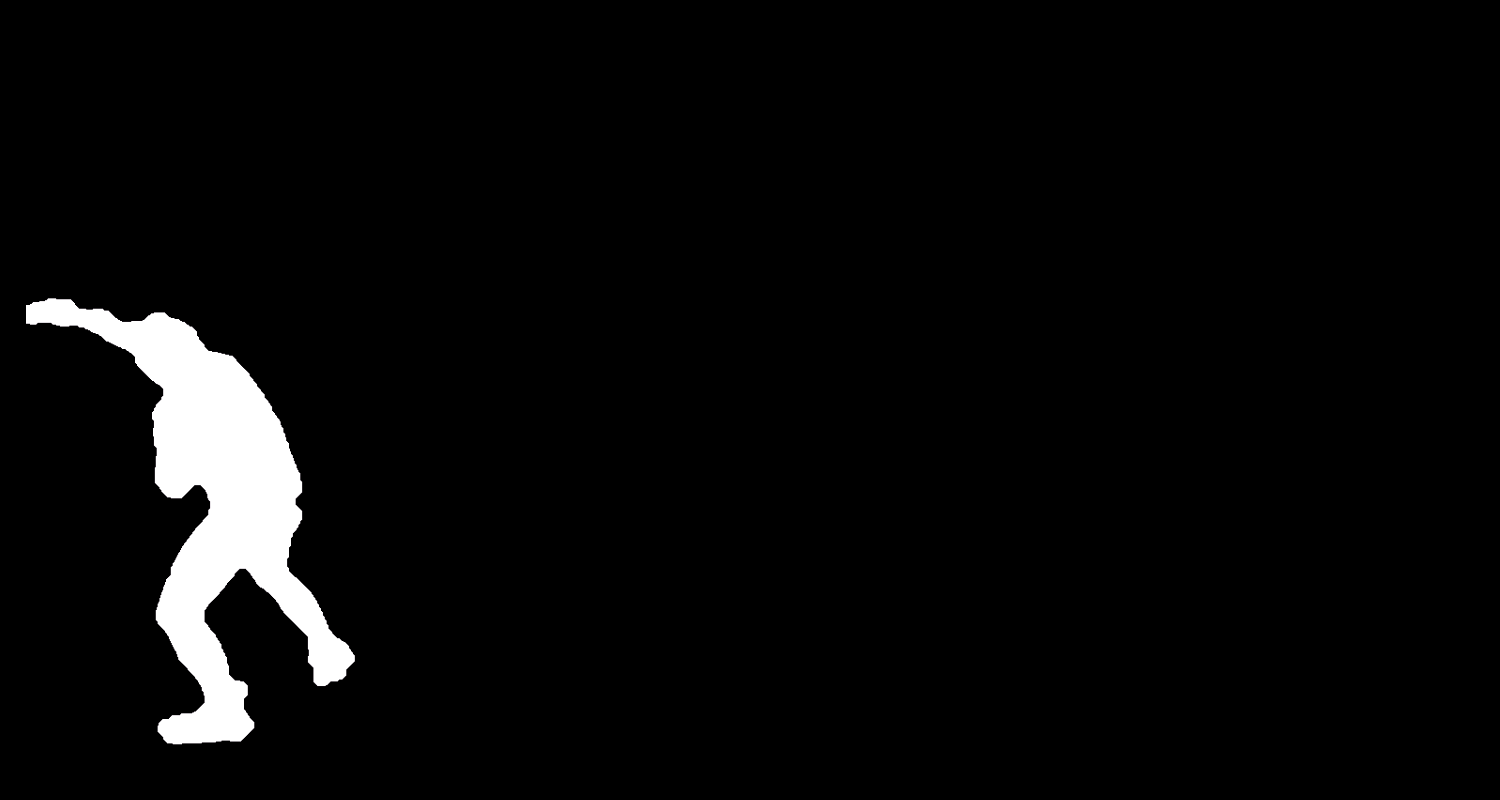}
        \captionsetup{labelformat=empty}
        \caption{F=400}
    \end{subfigure}
     \begin{subfigure}[b]{0.1\textwidth}
    \centering
        \includegraphics[width=\linewidth]{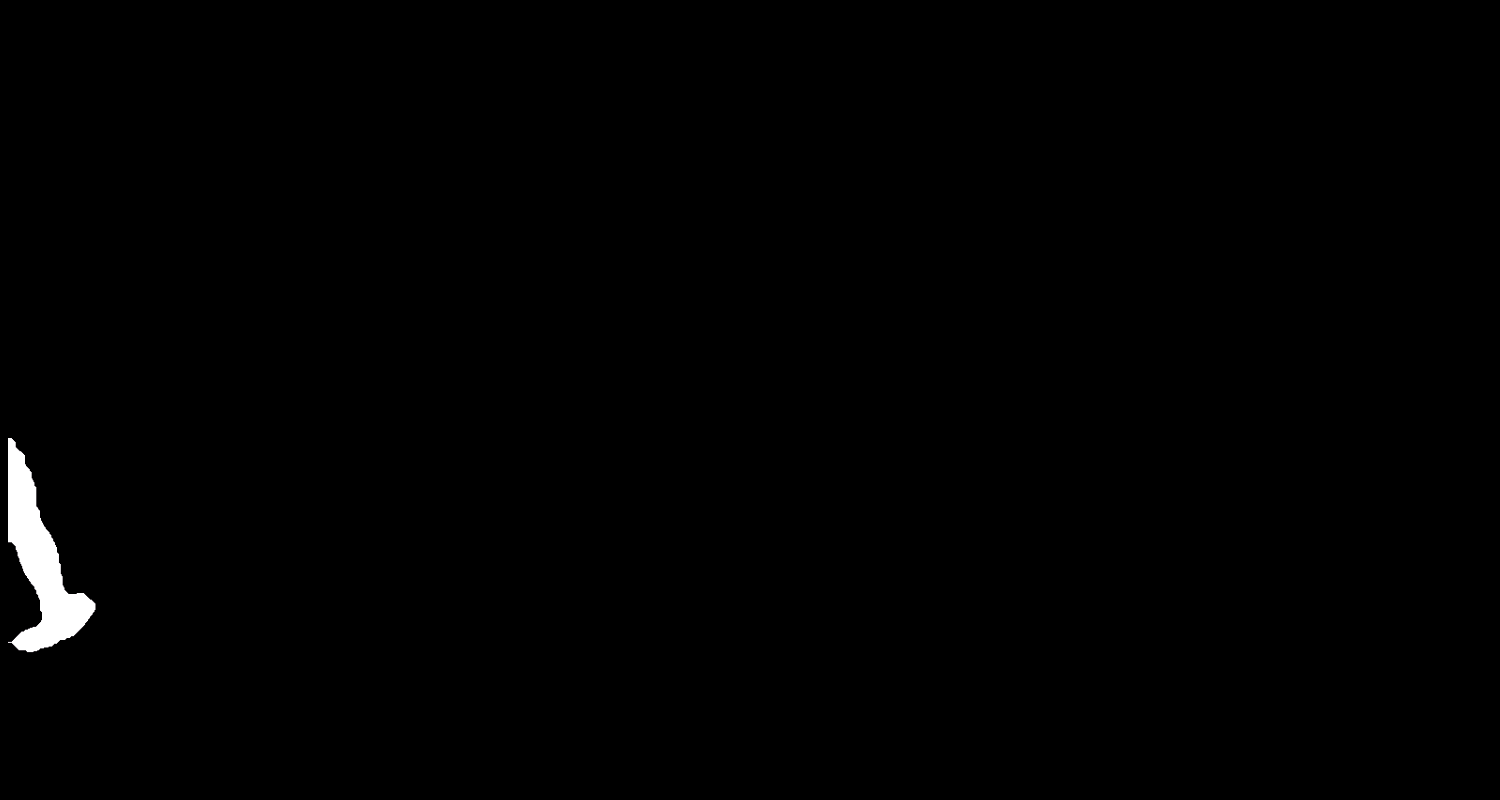}
        \captionsetup{labelformat=empty}
        \caption{F=600}
    \end{subfigure}
    \begin{subfigure}[b]{0.1\textwidth}
    \centering
        \includegraphics[width=\linewidth]{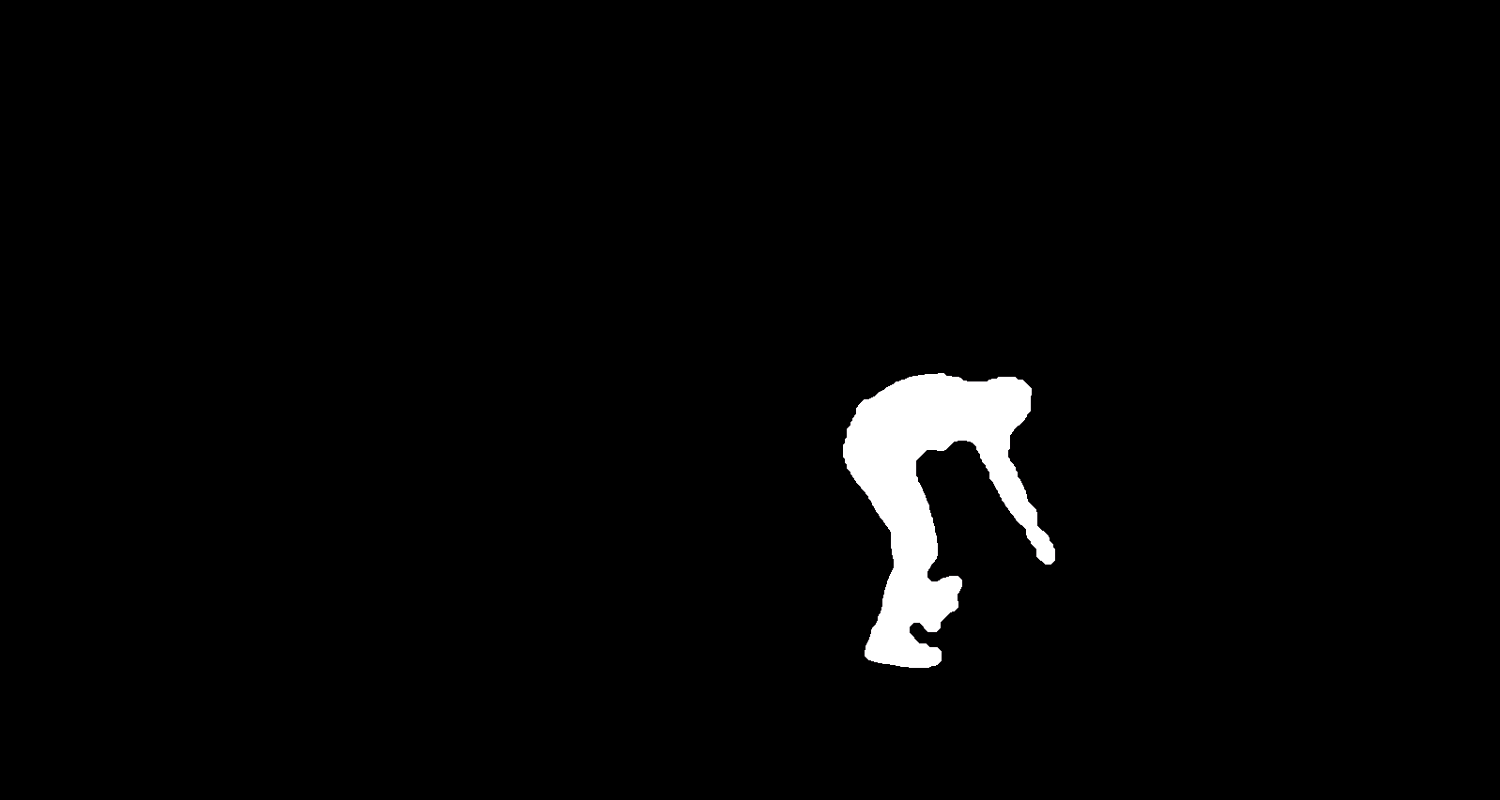}
        \captionsetup{labelformat=empty}
        \caption{F=1200}
    \end{subfigure}
    \begin{subfigure}[b]{0.1\textwidth}
    \centering
        \includegraphics[width=\linewidth]{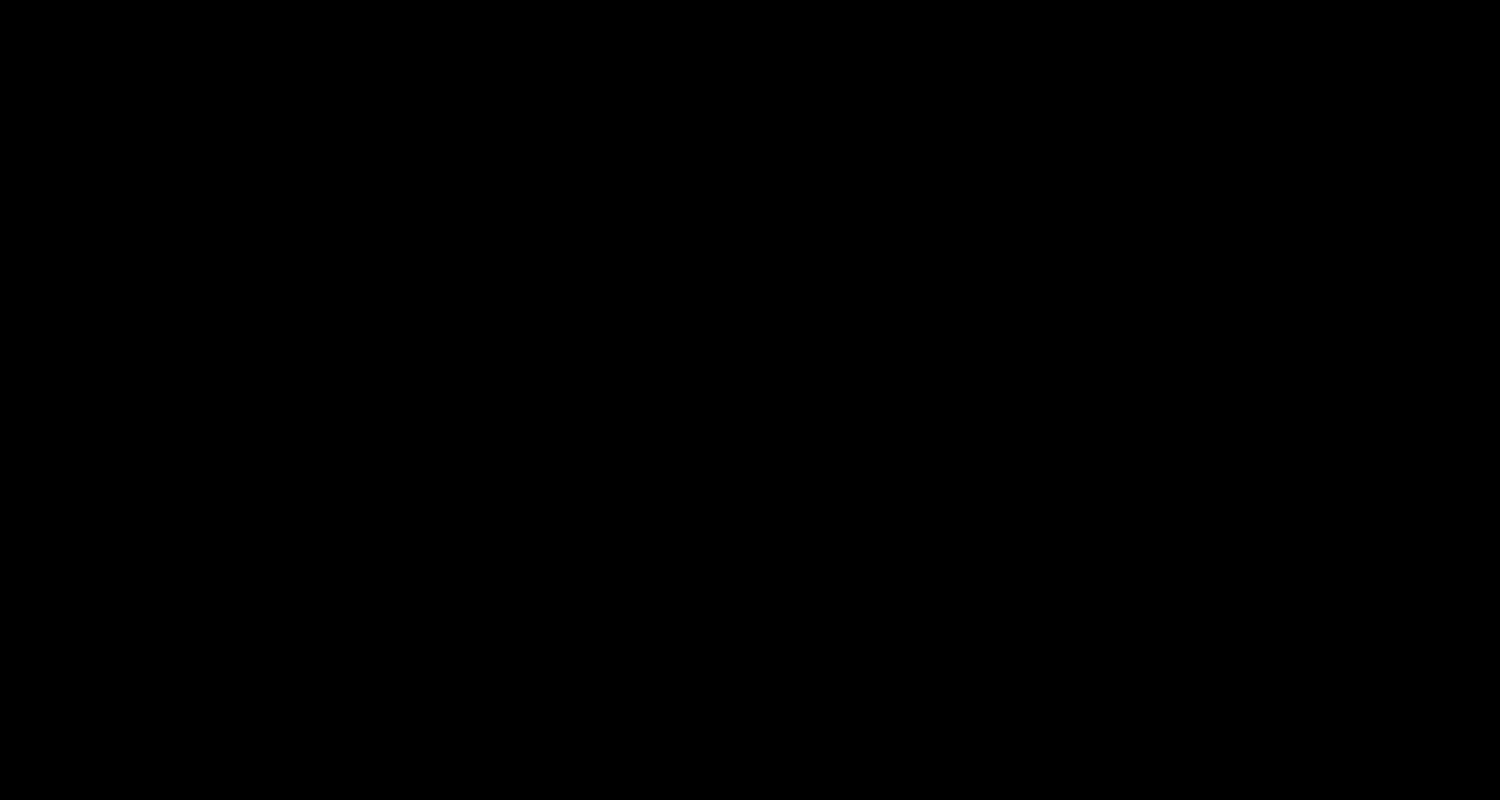}
        \captionsetup{labelformat=empty}
        \caption{F=1500}
    \end{subfigure}
\end{center}
    \vspace{-0.5cm}
   \caption{Per-frame mean joint error of test sequence S4 F3. The four matte images at bottom are captured by camera 4 at frame 400, 600, 1200 and 1500. Partially-captured frames, \textit{e.g.} frame 600 and 1500, show inferior performances. \textit{RS} for \textit{Random Shut}. \textit{F} denotes \textit{frame number}. (See \S\ref{sec:rs}).}
\label{fig:rs}
\vspace*{-14pt}
\end{figure}

\vspace*{-4pt}
\subsubsection{Human3.6M evaluation}
\label{sec:human_36}
\vspace*{-4pt}
 To validate the performance of our proposed multi-channel volume data representation and soft-argmax on volumetric data, we remove the IMU-bone layer of our network and test it on Human3.6M dataset \cite{h36m_pami}. It consists of 3.6M frames captured from 11 subjects with 4 synchronized cameras. Followed by the setting of \cite{sun2018integral}, protocol 1 uses subjects (S1, S5, S6, S7, S8, S9) for training and S11 for test. The measured result, mean per joint position error ( MPJPE ), is aligned by Procrustes Analysis ( PA MPJPE ). Protocol 2 uses subjects (S1, S5, S6, S7, S8) for training and subjects (S9, S11) for test without PA. To remove data redundancy, only every 5th frames in training sequences and every 64th frames in test sequences are used. Meanwhile, we use the provided foreground matte images from all the 4 cameras as input. Because the number of cameras is only 4 and subject always move within area captured by all cameras, Random Shut introduces more noise on this dataset in our experiment, which is not used in training as a result.
 
From table \ref{tab:human}, our method significantly outperforms other methods by large margins including singles-view methods and multi-view methods. Specifically, it improves the SOTA~\cite{tome2018rethinking} by \textbf{31.2mm} (relative 70\% lower than \cite{tome2018rethinking}) under protocol 1, and by \textbf{11.6mm} (relative 24.0\% lower than the state-of-the-art method \cite{kocabas2019self}) under protocol 2, and it also is clearly ahead by over
40\% as the same matte data used in \cite{trumble2018deep}.
By observing the failure cases, we find that the matte data of 3 actions, Greeting, Sitting Down and Waiting, are incomplete due to truncations, leading to unexpected high error. After remove these there actions, our result reaches 32.5mm. Furthermore, we will show the qualitative examples in the supplementary materials.

Finally, recent methods with volumetric input show overall better performance than that with 2D image input, showing the advantage of volumetric data representation. Also, our method outperforms other volume-based method by a large margin, showing the effectiveness of soft-argmax layer on volumetric data for 3D human pose estimation.

\vspace*{-4pt}
\subsection{Ablation study}
\vspace*{-4pt}
In this section, we try to answer the following three questions: (1) To what extent can IMU-bone layer contribute to the final estimation?  (2) Can the proposed data augmentation algorithm, Random Shut, improve the generalization capability of our model? And (3) To what extent and why 3D soft-argmax over volumetric data representation improve the estimation accuracy? All the experiments of ablation study are evaluated on the \textbf{TotalCapture dataset}\cite{Trumble:BMVC:2017}.

\vspace{-0.5cm}
\subsubsection{Sensor fusion}
\label{sec:fusion}

The first study explores the effectiveness of sensor fusion. Results of four training strategies are listed in Table~\ref{tab:fs}. To alleviate the possible influence of Random Shut (RS) on this ablation study, we perform comparison experiment between vision-IMU network and vision-only network \textit{w.} or \textit{w/o.} RS, respectively. Both of the vision-IMU networks outperform their vision-only counterparts, supporting the effectiveness of our data fusion solution quantitatively. 

In addition, we want to find out how IMU data influence the estimation after refinement. We plot the per-joint estimation error on test set. As shown in Fig~\ref{fig:pj}, the joints around limbs including foot and hand show more improvement by fusing IMU sensors compared to other joints. The main reason is that the IMU-bone layer constructs cylinder volume for each bone and it simulates volume around limbs much better than that around torso due to visual similarity. This argument is also supported by the qualitative result in Fig.\ref{fig:qual}. As can be observed, fusion result is superior to vision-only result. The joints around limbs show better refinement than that around torsp, especially under heavy self-occlusion.

As discussed before, the measurement of IMU sensors is more stable than that of vision sensors. Therefore, in supplementary material, we show that fusion approach shows better sequential stability than its vision-only counterpart.

\vspace{-0.2cm}
\subsubsection{Random Shut}
\vspace*{-4pt}
\label{sec:rs}

\begin{table}
\begin{center}
\begin{tabular}{|l|c|c|c|}
\hline
 Training & Full Frames & Partial Frames & Overall \\
\hline
with RS & $18.9$ & $34.5$ & $28.9$\\
without RS & $20.0$& $39.2$ & $32.4$\\
\hline
\end{tabular}
\end{center}
\vspace{-0.3cm}
\caption{Mean joint error of our proposed method trained \textit{w.} or \textit{w/o.} Random Shut (RS). Full frames are frames that are captured by all the 8 cameras while partial frames are not. All errors are measured in millimeter (mm) (See \S\ref{sec:rs}). }
\label{tab:rs}
\vspace*{-10pt}
\end{table}

The second study shows the motivation and effectiveness of the proposed data augmentation approach, Random Shut (RS), for multi-view data. Fig~\ref{fig:rs} demonstrates the per-frame estimation accuracy on a test sequence. We find that the estimation error increases when subject is partially-captured (Frame 600 and 1500), which is also proved by statistic data shown in Table~\ref{tab:rs}. Motivated by this finding, Random Shut simulates the situation when subject randomly missed by a certain camera during training. As demonstrated in Table~\ref{tab:rs} and in Fig.~\ref{fig:rs}, we conclude that RS shows improvement on generalization capability of the model for multi-view data, especially for the partially-captured frames.

\vspace{-0.3cm}
\subsubsection{Soft-argmax and volumetric representation}
\label{sec:soft_argmax}
\vspace*{-4pt}

\begin{figure}
\begin{center}
\includegraphics[width=\linewidth]{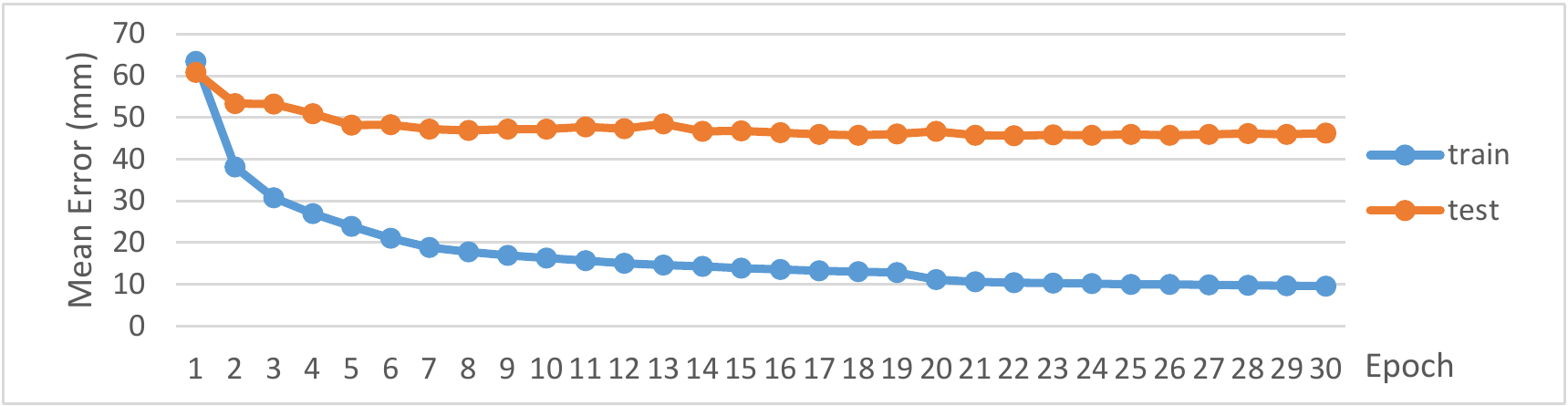}
\end{center}
    \vspace{-0.5cm}
  \caption{Mean error over epoch}
   \vspace{-0.2cm}
\label{fig:epoch}
\end{figure}

\label{sec:sa}
\begin{table}
\begin{center}
\resizebox{0.49\textwidth}{!}{
\setlength\tabcolsep{1pt}
\renewcommand\arraystretch{1.1}
\begin{tabular}{|l|c|c|c|}
\hline
Baselines & Hard-argmax & Soft-argmax &Direct regression (FC)  \\
\hline
 Mean Error (mm)& $39.7$& \textbf{32.7} & $45.2$ \\
\hline
\end{tabular}
}
\end{center}
\vspace{-0.3cm}
\caption{Mean joint error results for three kinds of output layers. FC stands for fully connected layer (See \S\ref{sec:soft_argmax}).}
\vspace{-0.6cm}
\label{tab:sf}
\end{table}

The last study aims to validate the effectiveness of 3D soft-argmax layer and volumetric representation. Although soft-argmax layer has already been used in human pose estimation by several works \cite{sun2018integral,luvizon20182d,nibali2018}, the effectiveness of this layer has not yet been evaluated in a fully 3D CNN-based network with volumetric input. 

To show the effectiveness of \textbf{3D soft-argmax}, we compare it with two widely used post-processing techniques: \textbf{hard-argmax} which directly picks highest voxel position, and \textbf{direct regression} which does regression via a dense layer. For fair comparison, we evaluate the three kinds of methods on the proposed fully 3D CNN-based network with volumetric data as input only. The comparative results in Table~\ref{tab:sf} show that soft-argmax significantly outperforms the other approaches. The main reason is that the low resolution of voxel heatmap limits the performance of hard-argmax as stated in section~\ref{sec:sal} and the dense layer is not well capable of learning argmax operator which is highly non-linear. 

To show the effectiveness of \textbf{volumetric representation}, we first have following finding in Table~\ref{tab:human} that our volumetric representation achieves mean error of \textbf{37.5mm}, an improvement of \textbf{23.6\%} and \textbf{28.9\%} compared to its 2D soft-argmax multiview counterparts 49.1mm\cite{Abdolrahim2018} and 52.8mm\cite{tome2018rethinking}, showing the advantage of volumetric representation in 3D human pose estimation. Second, our model converges very fast with 3D soft-argmax. As shown in Fig.~\ref{fig:epoch}, we achieve state-of-the-art result only after the first epoch, indicating that fully 3D CNN-based network would not waste its capability on learning unnecessary space mapping, which explains the effectiveness of volumetric representation. 

Meanwhile, the model converges too fast, indicating potential severe overfitting. We therefore explore how to tune the parameter $\theta$ in the soft-argmax Equation~\ref{equa:sam}, which may influence the convergence speed. As discussed in Sec.~\ref{sec:sal}, soft-argmax pays more attention to large voxel when $\theta$ enlarges. If $\theta$ is too large, soft-argmax will degenerate to hard-argmax. If $\theta$ is too small, too many voxels contribute to the final result, which may lead to overfitting. So we list some possible values of $\theta$ and their respective estimation errors in Table~\ref{tab:sfk}. In our experiment, $\theta=3$ achieves the best performance in our experiment setting. We believe the above findings of using soft-argmax in 3D space are novel and would be beneficial for future research.

\begin{table}[]
\begin{tabular}{|l|c|c|c|c|c|c|}
\hline
$\theta$        & 1    & 2    & 3   & 5    & 10   & 100  \\ \hline
Error (mm) & 33.8 & 33.4 & \textbf{32.7} & 34.2 & 38.9 & 41.0 \\ \hline
\end{tabular}
\caption{Different $\theta$ of soft-argmax and their respective estimation errors in the vision-only network (See \S\ref{sec:soft_argmax}).}
\label{tab:sfk}
\vspace*{-14pt}
\end{table}
\section{Limitation and future work}
\label{sec:limit}

One limitation is that, like many multi-view methods, the proposed multi-channel volume is biased to camera configuration. To make it fair in more general configurations, the triangulation should be trainable instead of fixed, making model adaptive. Another one is that performance of our method depends on the quality of foreground silhouette. The popular segmentation networks, like maskrcnn\cite{he2017mask}, show low quality of segmentation near human edgee so we use the given silhouette ground truth. In the future work, we will try to make this step into an end-to-end pipeline for simplifying the input representation and improve the silhouette quality simultaneously.
\section{Conclusion}
\label{sec:conc}
In this paper, we propose an IMU-aware network to fuse IMU data and multi-view images for 3D human pose estimation. To fully utilize the multi-view geometric information , we re-project it into a multi-channel volume format and apply Random Shut for data augmentation. To deeply fuse IMU orientation and multiple views, we then propose an IMU-bone layer to transform the original data from two modalities into a same feature space at early stage of network. Rigorous ablation shows the effectiveness of the multi-channel volume, Random Shut and IMU-bone layer. Finally, our method achieves the state-of-the-art performance on both TotalCapture and Human3.6M dataset with real-time capability. 


{\small
\bibliographystyle{ieee}
\bibliography{egbib}

\begin{thebibliography}{10}\itemsep=-1pt

\bibitem{bachmann1999orientation}
E.~R. Bachmann, I.~Duman, U.~Usta, R.~B. McGhee, X.~Yun, and M.~Zyda.
\newblock Orientation tracking for humans and robots using inertial sensors.
\newblock In {\em Proceedings 1999 IEEE International Symposium on
  Computational Intelligence in Robotics and Automation. CIRA'99 (Cat. No.
  99EX375)}, pages 187--194. IEEE, 1999.

\bibitem{bulat2016}
A.~Bulat and G.~Tzimiropoulos.
\newblock Human pose estimation via convolutional part heatmap regression.
\newblock In {\em European Conference on Computer Vision}, pages 717--732.
  Springer, 2016.

\bibitem{Carreira2016}
J.~Carreira, P.~Agrawal, K.~Fragkiadaki, and J.~Malik.
\newblock Human pose estimation with iterative error feedback.
\newblock {\em Proceedings of the IEEE Conference on Computer Vision and
  Pattern Recognition}, pages 4733--4742, 2016.

\bibitem{chen2017}
C.-H. Chen and D.~Ramanan.
\newblock 3d human pose estimation= 2d pose estimation+ matching.
\newblock In {\em Proceedings of the IEEE Conference on Computer Vision and
  Pattern Recognition}, pages 7035--7043, 2017.

\bibitem{chen2019weakly}
X.~Chen, K.-Y. Lin, W.~Liu, C.~Qian, and L.~Lin.
\newblock Weakly-supervised discovery of geometry-aware representation for 3d
  human pose estimation.
\newblock In {\em Proceedings of the IEEE Conference on Computer Vision and
  Pattern Recognition}, pages 10895--10904, 2019.

\bibitem{Chen2014}
Y.~A. Chen, X.
\newblock Articulated pose estimation by a graphical model with image dependent
  pairwise relations.
\newblock {\em In the Neural Information Processing Systems}, page 3041–3048,
  2014.

\bibitem{choukroun2003novel}
D.~Choukroun.
\newblock {\em Novel methods for attitude determination using vector
  observations}.
\newblock Technion-Israel Institute of Technology, Faculty of Aerospace
  Engineering, 2003.

\bibitem{chu2017}
X.~Chu, W.~Yang, W.~Ouyang, C.~Ma, A.~L. Yuille, and X.~Wang.
\newblock Multi-context attention for human pose estimation.
\newblock {\em Proceedings of the IEEE Conference on Computer Vision and
  Pattern Recognition}, pages 1831--1840, 2017.

\bibitem{dantone2013human}
M.~Dantone, J.~Gall, C.~Leistner, and L.~Van~Gool.
\newblock Human pose estimation using body parts dependent joint regressors.
\newblock In {\em Proceedings of the IEEE Conference on Computer Vision and
  Pattern Recognition}, pages 3041--3048, 2013.

\bibitem{dong2019fast}
J.~Dong, W.~Jiang, Q.~Huang, H.~Bao, and X.~Zhou.
\newblock Fast and robust multi-person 3d pose estimation from multiple views.
\newblock In {\em Proceedings of the IEEE Conference on Computer Vision and
  Pattern Recognition}, pages 7792--7801, 2019.

\bibitem{fang2018learning}
H.-S. Fang, Y.~Xu, W.~Wang, X.~Liu, and S.-C. Zhu.
\newblock Learning pose grammar to encode human body configuration for 3d pose
  estimation.
\newblock In {\em Thirty-Second AAAI Conference on Artificial Intelligence},
  2018.

\bibitem{gebre2004design}
D.~Gebre-Egziabher, R.~C. Hayward, and J.~D. Powell.
\newblock Design of multi-sensor attitude determination systems.
\newblock {\em IEEE Transactions on aerospace and electronic systems},
  40(2):627--649, 2004.

\bibitem{he2017mask}
K.~He, G.~Gkioxari, P.~Doll{\'a}r, and R.~Girshick.
\newblock Mask r-cnn.
\newblock In {\em Proceedings of the IEEE international conference on computer
  vision}, pages 2961--2969, 2017.

\bibitem{huang2018}
F.~Huang, A.~Zeng, M.~Liu, J.~Qin, and Q.~Xu.
\newblock Structure-aware 3d hourglass network for hand pose estimation from
  single depth image.
\newblock In {\em British Machine Vision Conference}, page 289, 2018.

\bibitem{deepercut1}
E.~Insafutdinov, L.~Pishchulin, B.~Andres, M.~Andriluka, and B.~Schiele.
\newblock Deepercut: A deeper, stronger, and faster multi-person pose
  estimation model.
\newblock In {\em European Conference on Computer Vision}, pages 34--50.
  Springer, 2016.

\bibitem{h36m_pami}
C.~Ionescu, D.~Papava, V.~Olaru, and C.~Sminchisescu.
\newblock Human3.6m: Large scale datasets and predictive methods for 3d human
  sensing in natural environments.
\newblock {\em IEEE Transactions on Pattern Analysis and Machine Intelligence},
  2014.

\bibitem{Abdolrahim2018}
A.~Kadkhodamohammadi and N.~Padoy.
\newblock A generalizable approach for multi-view 3d human pose regression.
\newblock {\em CoRR}, abs/1804.10462, 2018.

\bibitem{Ke2018}
L.~Ke, M.-C. Chang, H.~Qi, and S.~Lyu.
\newblock Multi-scale structure-aware network for human pose estimation.
\newblock In {\em The European Conference on Computer Vision (ECCV)}, September
  2018.

\bibitem{kocabas2019self}
M.~Kocabas, S.~Karagoz, and E.~Akbas.
\newblock Self-supervised learning of 3d human pose using multi-view geometry.
\newblock {\em arXiv preprint arXiv:1903.02330}, 2019.

\bibitem{li2014}
S.~Li and A.~B. Chan.
\newblock 3d human pose estimation from monocular images with deep
  convolutional neural network.
\newblock In {\em Asian Conference on Computer Vision}, pages 332--347.
  Springer, 2014.

\bibitem{lin2017}
M.~Lin, L.~Lin, X.~Liang, K.~Wang, and H.~Cheng.
\newblock Recurrent 3d pose sequence machines.
\newblock In {\em Computer Vision and Pattern Recognition (CVPR), 2017 IEEE
  Conference on}, pages 5543--5552. IEEE, 2017.

\bibitem{loper2015smpl}
M.~Loper, N.~Mahmood, J.~Romero, G.~Pons-Moll, and M.~J. Black.
\newblock Smpl: A skinned multi-person linear model.
\newblock {\em ACM transactions on graphics (TOG)}, 34(6):248, 2015.

\bibitem{luvizon20182d}
D.~C. Luvizon, D.~Picard, and H.~Tabia.
\newblock 2d/3d pose estimation and action recognition using multitask deep
  learning.
\newblock In {\em Proceedings of the IEEE Conference on Computer Vision and
  Pattern Recognition}, pages 5137--5146, 2018.

\bibitem{luvizon2017human}
D.~C. Luvizon, H.~Tabia, and D.~Picard.
\newblock Human pose regression by combining indirect part detection and
  contextual information.
\newblock {\em arXiv preprint arXiv:1710.02322}, 2017.

\bibitem{8259006}
M.~{Ma}, C.~{Sun}, and X.~{Chen}.
\newblock Deep coupling autoencoder for fault diagnosis with multimodal sensory
  data.
\newblock {\em IEEE Transactions on Industrial Informatics}, 14(3):1137--1145,
  March 2018.

\bibitem{malleson2017real}
C.~Malleson, A.~Gilbert, M.~Trumble, J.~Collomosse, A.~Hilton, and M.~Volino.
\newblock Real-time full-body motion capture from video and imus.
\newblock In {\em 3D Vision (3DV), 2017 International Conference on}, pages
  449--457. IEEE, 2017.

\bibitem{martinez2017}
J.~Martinez, R.~Hossain, J.~Romero, and J.~J. Little.
\newblock A simple yet effective baseline for 3d human pose estimation.
\newblock In {\em International Conference on Computer Vision}, volume~1,
  page~5, 2017.

\bibitem{mehta2017}
D.~Mehta, S.~Sridhar, O.~Sotnychenko, H.~Rhodin, M.~Shafiei, H.-P. Seidel,
  W.~Xu, D.~Casas, and C.~Theobalt.
\newblock Vnect: Real-time 3d human pose estimation with a single rgb camera.
\newblock {\em ACM Transactions on Graphics (TOG)}, 36(4):44, 2017.

\bibitem{newell2016stacked}
A.~Newell, K.~Yang, and J.~Deng.
\newblock Stacked hourglass networks for human pose estimation.
\newblock In {\em European Conference on Computer Vision}, pages 483--499.
  Springer, 2016.

\bibitem{nibali2018}
A.~Nibali, Z.~He, S.~Morgan, and L.~Prendergast.
\newblock 3d human pose estimation with 2d marginal heatmaps.
\newblock {\em arXiv preprint arXiv:1806.01484}, 2018.

\bibitem{nibali2018nu}
A.~Nibali, Z.~He, S.~Morgan, and L.~Prendergast.
\newblock Numerical coordinate regression with convolutional neural networks.
\newblock {\em arXiv preprint arXiv:1801.07372}, 2018.

\bibitem{papandreou2017towards}
G.~Papandreou, T.~Zhu, N.~Kanazawa, A.~Toshev, J.~Tompson, C.~Bregler, and
  K.~Murphy.
\newblock Towards accurate multi-person pose estimation in the wild.
\newblock In {\em Proceedings of the IEEE Conference on Computer Vision and
  Pattern Recognition}, pages 4903--4911, 2017.

\bibitem{paulichxsens}
M.~Paulich, M.~Schepers, N.~Rudigkeit, and G.~Bellusci.
\newblock Xsens mtw awinda: Miniature wireless inertial-magnetic motion tracker
  for highly accurate 3d kinematic applications.

\bibitem{Pavlakos2016}
G.~Pavlakos, X.~Zhou, K.~G. Derpanis, and K.~Daniilidis.
\newblock Coarse-to-fine volumetric prediction for single-image 3d human pose.
\newblock {\em IEEE Conference on Computer Vision and Pattern Recognition},
  abs/1611.07828, 2017.

\bibitem{PavlakosZDD17}
G.~Pavlakos, X.~Zhou, K.~G. Derpanis, and K.~Daniilidis.
\newblock Harvesting multiple views for marker-less 3d human pose annotations.
\newblock {\em Conference on Computer Vision and Pattern Recognition}, page
  1253–1262, 2017.

\bibitem{Pons11}
Pons-Moll, G.~Baak, A., Gall, J., Leal-Taixe, L., Muller, M., Seidel, H.P.,
  Rosenhahn, and B.
\newblock Outdoor human motion capture using inverse kinematics and von
  mises-fisher sampling, 2011.

\bibitem{rhodin2018learning}
H.~Rhodin, J.~Sp{\"o}rri, I.~Katircioglu, V.~Constantin, F.~Meyer,
  E.~M{\"u}ller, M.~Salzmann, and P.~Fua.
\newblock Learning monocular 3d human pose estimation from multi-view images.
\newblock In {\em Proceedings of the IEEE Conference on Computer Vision and
  Pattern Recognition}, pages 8437--8446, 2018.

\bibitem{Slyper08}
R.~Slyper, Hodgins, and J.K.
\newblock Action capture with accelerometers.
\newblock {\em In Proceedings of the 2008 ACM SIGGRAPH/Eurographics Symposium
  on Computer Animation}, pages 193--199, 2008.

\bibitem{sun2018integral}
X.~Sun, B.~Xiao, F.~Wei, S.~Liang, and Y.~Wei.
\newblock Integral human pose regression.
\newblock In {\em Proceedings of the European Conference on Computer Vision
  (ECCV)}, pages 529--545, 2018.

\bibitem{srndi2018}
I.~Sárándi, T.~Linder, K.~O. Arras, and B.~Leibe.
\newblock Synthetic occlusion augmentation with volumetric heatmaps for the
  2018 eccv posetrack challenge on 3d human pose estimation, 2018.

\bibitem{Tautges11}
Tautges, J., Zinke, A.~Krüger, B., Baumann, J., Weber, A., Helten, T.,
  Müller, M., Seidel, H.P., and B.~Eberhardt.
\newblock Motion reconstruction using sparse accelerometer data.
\newblock {\em ACM transactions on graphics (TOG)}, 30(3):18, 2011.

\bibitem{tekin2017}
B.~Tekin, P.~Marquez~Neila, M.~Salzmann, and P.~Fua.
\newblock Learning to fuse 2d and 3d image cues for monocular body pose
  estimation.
\newblock In {\em International Conference on Computer Vision (ICCV)}, number
  EPFL-CONF-230311, 2017.

\bibitem{tome2018rethinking}
D.~Tome, M.~Toso, L.~Agapito, and C.~Russell.
\newblock Rethinking pose in 3d: Multi-stage refinement and recovery for
  markerless motion capture.
\newblock In {\em 2018 International Conference on 3D Vision (3DV)}, pages
  474--483. IEEE, 2018.

\bibitem{tompson2015efficient}
J.~Tompson, R.~Goroshin, A.~Jain, Y.~LeCun, and C.~Bregler.
\newblock Efficient object localization using convolutional networks.
\newblock In {\em Proceedings of the IEEE Conference on Computer Vision and
  Pattern Recognition}, pages 648--656, 2015.

\bibitem{tompson2014joint}
J.~J. Tompson, A.~Jain, Y.~LeCun, and C.~Bregler.
\newblock Joint training of a convolutional network and a graphical model for
  human pose estimation.
\newblock In {\em Advances in neural information processing systems}, pages
  1799--1807, 2014.

\bibitem{trumble2018deep}
M.~Trumble, A.~Gilbert, A.~Hilton, and J.~Collomosse.
\newblock Deep autoencoder for combined human pose estimation and body model
  upscaling.
\newblock In {\em European conference on computer vision (ECCV’18)}, 2018.

\bibitem{trumble2017total}
M.~Trumble, A.~Gilbert, C.~Malleson, A.~Hilton, and J.~Collomosse.
\newblock Total capture: 3d human pose estimation fusing video and inertial
  sensors.
\newblock In {\em Proceedings of 28th British Machine Vision Conference}, pages
  1--13, 2017.

\bibitem{Trumble:BMVC:2017}
M.~Trumble, A.~Gilbert, C.~Malleson, A.~Hilton, and J.~Collomosse.
\newblock Total capture: 3d human pose estimation fusing video and inertial
  sensors.
\newblock In {\em 2017 British Machine Vision Conference (BMVC)}, 2017.

\bibitem{Vicon2014:Online}
Vicon.
\newblock Vicon motion systems ltd.
\newblock http://www.vicon.com/.

\bibitem{Vlasic07}
D.~Vlasic, R.~Adelsberger, G.~Vannucci, J.~Barnwell, M.~Gross, W.~Matusik, and
  J.~Popović.
\newblock Practical motion capture in everyday surroundings.
\newblock {\em ACM transactions on graphics (TOG)}, 26(3):35, 2007.

\bibitem{von2018recovering}
T.~von Marcard, R.~Henschel, M.~J. Black, B.~Rosenhahn, and G.~Pons-Moll.
\newblock Recovering accurate 3d human pose in the wild using imus and a moving
  camera.
\newblock In {\em European Conference on Computer Vision (ECCV)}, 2018.

\bibitem{von2016human}
T.~von Marcard, G.~Pons-Moll, and B.~Rosenhahn.
\newblock Human pose estimation from video and imus.
\newblock {\em IEEE transactions on pattern analysis and machine intelligence},
  38(8):1533--1547, 2016.

\bibitem{von2017sparse}
T.~von Marcard, B.~Rosenhahn, M.~J. Black, and G.~Pons-Moll.
\newblock Sparse inertial poser: Automatic 3d human pose estimation from sparse
  imus.
\newblock In {\em Computer Graphics Forum}, volume~36, pages 349--360. Wiley
  Online Library, 2017.

\bibitem{wang2014}
C.~Wang, Y.~Wang, Z.~Lin, A.~L. Yuille, and W.~Gao.
\newblock Robust estimation of 3d human poses from a single image.
\newblock In {\em Proceedings of the IEEE Conference on Computer Vision and
  Pattern Recognition}, pages 2361--2368, 2014.

\bibitem{wei2016convolutional}
S.-E. Wei, V.~Ramakrishna, T.~Kanade, and Y.~Sheikh.
\newblock Convolutional pose machines.
\newblock In {\em Proceedings of the IEEE Conference on Computer Vision and
  Pattern Recognition}, pages 4724--4732, 2016.

\bibitem{Xsens2009:Online}
Xsens.
\newblock Xsens motion technologies.
\newblock http://www.xsens.com/.

\bibitem{yang2017}
W.~Yang, S.~Li, W.~Ouyang, H.~Li, and X.~Wang.
\newblock Learning feature pyramids for human pose estimation.
\newblock In {\em The IEEE International Conference on Computer Vision (ICCV)},
  volume~2, 2017.

\bibitem{yasin2016}
H.~Yasin, U.~Iqbal, B.~Kruger, A.~Weber, and J.~Gall.
\newblock A dual-source approach for 3d pose estimation from a single image.
\newblock In {\em Proceedings of the IEEE Conference on Computer Vision and
  Pattern Recognition}, pages 4948--4956, 2016.

\end{thebibliography}
}

\end{document}